\documentclass[twoside]{article}

\usepackage[accepted]{aistats2022}
\usepackage[round]{natbib}

\usepackage{url}
\usepackage{subfig}
\usepackage{multirow}
\usepackage{wrapfig}
\usepackage{enumerate}
\usepackage{booktabs}
\usepackage{mathtools}
\usepackage[makeroom]{cancel}
\usepackage[title]{appendix}
\usepackage{moresize}
\usepackage{todonotes}
\usepackage{lipsum}
\usepackage{siunitx}
\usepackage{soul}
\usepackage{rotating}
\usepackage{layouts}
\usepackage{colortbl}

\usepackage{amsmath,amsfonts,bm}

\usepackage{amsthm}

\theoremstyle{definition}

\def\1{\bm{1}}

\DeclareMathAlphabet{\mathsfit}{\encodingdefault}{\sfdefault}{m}{sl}
\SetMathAlphabet{\mathsfit}{bold}{\encodingdefault}{\sfdefault}{bx}{n}

\newcommand{\E}{\mathbb{E}}

\newcommand{\R}{\mathbb{R}}

\DeclareMathOperator*{\argmax}{arg\,max}

\renewcommand{\R}{\mathbb{R}}
\renewcommand{\E}{\mathop{\mathbb{E}}}
\newcommand{\D}{\mathcal{D}}

\newcommand{\Dout}{\mathcal{D}_\text{out}}

\newcommand{\xout}{x_\text{out}}

\newcommand{\abs}[1]{\vert #1 \vert}

\usepackage{thmtools}
\usepackage{thm-restate}

\colorlet{tablegray}{gray!30}
\colorlet{hlpurple}{purple!20}
\sethlcolor{hlpurple}

\usepackage{pgfplots}
\pgfplotsset{compat=newest}
\pgfplotsset{scaled y ticks=false}
\usepgfplotslibrary{groupplots}
\usepgfplotslibrary{dateplot}

\graphicspath{{figs/}}

\usepackage{tikz}
\usetikzlibrary{external}
\usepackage{hyperref}
\definecolor{mydarkblue}{rgb}{0,0.08,0.45}
\hypersetup{ %
  colorlinks=true,
  linkcolor=mydarkblue,
  citecolor=mydarkblue,
  filecolor=mydarkblue,
  urlcolor=mydarkblue,
}

\usepackage[capitalise]{cleveref}
\Crefname{appsec}{Appendix}{Appendices}

\usepackage{algorithm}
\usepackage{algpseudocode}

\newcolumntype{H}{>{\setbox0=\hbox\bgroup}c<{\egroup}@{}}

\begin{document}

  %

  %

  \twocolumn[

  \aistatstitle{Being a Bit Frequentist Improves Bayesian Neural Networks}

  \aistatsauthor{ Agustinus Kristiadi \And Matthias Hein \And Philipp Hennig }

  \aistatsaddress{University of T\"{u}bingen \And  University of T\"{u}bingen \And University of T\"{u}bingen \\ MPI for Intelligent Systems, T\"{u}bingen} ]

  \begin{abstract}
    Despite their compelling theoretical properties, Bayesian neural networks (BNNs) tend to perform worse than frequentist methods in classification-based uncertainty quantification (UQ) tasks such as out-of-distribution (OOD) detection.
    In this paper, based on empirical findings in prior works, we hypothesize that this issue is because even recent Bayesian methods have never considered OOD data in their training processes, even though this ``OOD training'' technique is an integral part of state-of-the-art frequentist UQ methods.
    To validate this, we treat OOD data as a first-class citizen in BNN training by exploring four different ways of incorporating OOD data into Bayesian inference.
    We show in extensive experiments that OOD-trained BNNs are competitive to recent frequentist baselines.
    This work thus provides strong baselines for future work in Bayesian UQ.
  \end{abstract}

  \section{INTRODUCTION}\label{sec:intro}

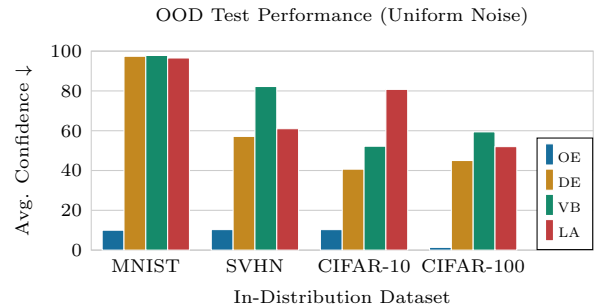
\begin{figure}
  \def\figonewidth{1}
  \def\figoneheight{0.18}

  \subfloat{

\begin{tikzpicture}[baseline]

\definecolor{color0}{rgb}{0.0906862745098039,0.425980392156863,0.611274509803922}
\definecolor{color1}{rgb}{0.76421568627451,0.531862745098039,0.125980392156863}
\definecolor{color2}{rgb}{0.084313725490196,0.543137254901961,0.416666666666667}
\definecolor{color3}{rgb}{0.75343137254902,0.238725490196078,0.241666666666667}

\tikzstyle{every node}=[font=\scriptsize]

\pgfplotsset{compat=1.11,
  /pgfplots/ybar legend/.style={
  /pgfplots/legend image code/.code={%
   \draw[##1,/tikz/.cd,yshift=-0.25em]
    (0cm,0cm) rectangle (3pt,0.8em);},
 },
}

\begin{axis}[
width=\figonewidth\linewidth,
height=\figoneheight\textheight,
legend style={nodes={scale=0.75, transform shape}, at={(1,0)}, anchor=south east, draw=black},
title={OOD Test Performance (Uniform Noise)},
axis line style={white!80!black},
xtick style={draw=none},
ytick style={draw=none},
x grid style={white!80!black},
xlabel={In-Distribution Dataset},
ylabel={Avg. Confidence \(\downarrow\)},
xmajorticks=true,
xmin=-0.5, xmax=4.1,
xtick style={draw=none},
xtick={0,1,2,3},
xticklabels={MNIST,SVHN,CIFAR-10,CIFAR-100},
y grid style={white!80!black},
ymajorgrids=true,
ymajorticks=true,
ymin=0, ymax=100,
ytick style={draw=none}
]
\draw[draw=white,fill=color0] (axis cs:-0.4,0) rectangle (axis cs:-0.2,10.0050024688244);
\addlegendimage{ybar,ybar legend,draw=white,fill=color0};
\addlegendentry{OE}

\draw[draw=white,fill=color0] (axis cs:0.6,0) rectangle (axis cs:0.8,10.3104636073112);
\draw[draw=white,fill=color0] (axis cs:1.6,0) rectangle (axis cs:1.8,10.2972812950611);
\draw[draw=white,fill=color0] (axis cs:2.6,0) rectangle (axis cs:2.8,1.3769936747849);
\draw[draw=white,fill=color1] (axis cs:-0.2,0) rectangle (axis cs:0,97.421008348465);
\addlegendimage{ybar,ybar legend,draw=white,fill=color1};
\addlegendentry{DE}

\draw[draw=white,fill=color1] (axis cs:0.8,0) rectangle (axis cs:1,57.147741317749);
\draw[draw=white,fill=color1] (axis cs:1.8,0) rectangle (axis cs:2,40.6866252422333);
\draw[draw=white,fill=color1] (axis cs:2.8,0) rectangle (axis cs:3,45.0297355651855);
\draw[draw=white,fill=color2] (axis cs:2.77555756156289e-17,0) rectangle (axis cs:0.2,97.8235065937042);
\addlegendimage{ybar,ybar legend,draw=white,fill=color2};
\addlegendentry{VB}

\draw[draw=white,fill=color2] (axis cs:1,0) rectangle (axis cs:1.2,82.2238123416901);
\draw[draw=white,fill=color2] (axis cs:2,0) rectangle (axis cs:2.2,52.2113478183746);
\draw[draw=white,fill=color2] (axis cs:3,0) rectangle (axis cs:3.2,59.4188570976257);
\draw[draw=white,fill=color3] (axis cs:0.2,0) rectangle (axis cs:0.4,96.5552830696106);
\addlegendimage{ybar,ybar legend,draw=white,fill=color3};
\addlegendentry{LA}

\draw[draw=white,fill=color3] (axis cs:1.2,0) rectangle (axis cs:1.4,61.0107612609863);
\draw[draw=white,fill=color3] (axis cs:2.2,0) rectangle (axis cs:2.4,80.7773649692535);
\draw[draw=white,fill=color3] (axis cs:3.2,0) rectangle (axis cs:3.4,52.0085990428925);
\addplot [line width=1.08pt, white!26.0!black]
table {%
-0.3 nan
-0.3 nan
};
\addplot [line width=1.08pt, white!26.0!black]
table {%
0.7 nan
0.7 nan
};
\addplot [line width=1.08pt, white!26.0!black]
table {%
1.7 nan
1.7 nan
};
\addplot [line width=1.08pt, white!26.0!black]
table {%
2.7 nan
2.7 nan
};
\addplot [line width=1.08pt, white!26.0!black]
table {%
-0.1 nan
-0.1 nan
};
\addplot [line width=1.08pt, white!26.0!black]
table {%
0.9 nan
0.9 nan
};
\addplot [line width=1.08pt, white!26.0!black]
table {%
1.9 nan
1.9 nan
};
\addplot [line width=1.08pt, white!26.0!black]
table {%
2.9 nan
2.9 nan
};
\addplot [line width=1.08pt, white!26.0!black]
table {%
0.1 nan
0.1 nan
};
\addplot [line width=1.08pt, white!26.0!black]
table {%
1.1 nan
1.1 nan
};
\addplot [line width=1.08pt, white!26.0!black]
table {%
2.1 nan
2.1 nan
};
\addplot [line width=1.08pt, white!26.0!black]
table {%
3.1 nan
3.1 nan
};
\addplot [line width=1.08pt, white!26.0!black]
table {%
0.3 nan
0.3 nan
};
\addplot [line width=1.08pt, white!26.0!black]
table {%
1.3 nan
1.3 nan
};
\addplot [line width=1.08pt, white!26.0!black]
table {%
2.3 nan
2.3 nan
};
\addplot [line width=1.08pt, white!26.0!black]
table {%
3.3 nan
3.3 nan
};
\end{axis}

\end{tikzpicture}}

  \caption{
    Average confidence on uniform OOD test data (lower is better).
    LA, VB, DE, and OE stand for the Laplace approximation, variational Bayes, Deep Ensemble, and Outlier Exposure, respectively.
    All methods have similar accuracy and confidence on all the in-distribution test sets, but OE is the only method that achieves low confidence on these outliers.
  }
  \label{fig:one}
\end{figure}

Uncertainty quantification (UQ) allows learning systems to ``know when they do not know''.
It is important functionality, especially in safety-critical applications of AI \citep{amodei2016concrete}, where a system encountering a novel task (which should be associated with high uncertainty) should hesitate or notify a human supervisor.
Both the Bayesian and frequentist deep learning communities address similar UQ functionality, but it appears that even recent Bayesian neural networks \citep[BNNs,][etc.]{osawa2019practical,tomczak2020efficient,dusenberry2020rankone,izmailov2021dangers,izmailov2021hmc} tend to underperform compared to the state-of-the-art frequentist UQ methods \citep[etc.]{hendrycks2018deep,lee2018training,hein2019relu,meinke2020towards,bitterwolf2020certifiably}.
\Cref{fig:one} shows this observation: the frequentist Outlier Exposure method \citep{hendrycks2018deep} performs much better than BNNs and even Deep Ensemble \citep{lakshminarayanan2017simple}, which has been considered as a strong baseline in Bayesian deep learning \citep{ovadia2019can}.

This paper thus seeks to answer the question of ``how can we bring the performance of BNNs on par with that of recent frequentist UQ methods?''
Our working hypothesis is that the disparity between them is \emph{not} due to some fundamental advantage of the frequentist viewpoint.
Rather, it is due to the more mundane practical fact that recent frequentist UQ methods leverage OOD data in their training process, via the so-called ``OOD training'' technique.
The benefits of this technique are well-studied, both for improving generalization \citep{zhang2017universum} and more recently, for OOD detection \citep{lee2018training,hein2019relu,meinke2020towards,bitterwolf2020certifiably}.
But while OOD data have been used for tuning the hyperparameters of BNNs \citep{kristiadi2020being}, it appears that even recently proposed deep Bayesian methods have not considered OOD training.
A reason for this may be that it is unclear how one can incorporate OOD data in the Bayesian inference itself.

We explore four options---some are philosophically clean, others are heuristics---of incorporating OOD data into Bayesian inference.
These methods are motivated by the assumptions that the data (i) have an extra ``none class'', (ii) are entirely represented by ``soft labels'', or (iii) have mixed ``hard'' and ``soft labels''.
Moreover, we also (iv) investigate an interpretation of the popular OE loss as a likelihood that can readily be used in Bayesian inference.
We compare the four of them against strong baselines in various UQ tasks and show that BNNs equipped with these likelihoods can outperform a recent OOD-trained frequentist baseline.
Our empirical findings thus validate the hypothesis that OOD training is the cause of BNNs underperformance.
We hope that the proposed approaches, especially the simple yet best-performing ``none-class'' method, can serve as strong baselines in the Bayesian deep learning community.

  \section{PRELIMINARIES}\label{sec:prelim}

We focus on classification tasks.
Let \(F: \R^n \times \R^d \to \R^c\) defined by \((x, \theta) \mapsto F(x; \theta)\) be an \(\ell\)-layer, \(c\)-class NN with any activation function.
Here, \(\R^n\), \(\R^d\), and \(\R^c\) are the input, parameter, and output spaces of the network, respectively.
Let \(P(X)\) and \(P(Y | X)\) be unknown probability distributions on \(\R^n\) and \(\{ 1, \dots, c \}\), respectively.
Given an i.i.d.\ dataset \(\D := \{ (x^{(i)}, y^{(i)}) \}_{i=1}^m\) sampled from the previous distributions, the standard training procedure for $F$ amounts (from the Bayesian perspective) to \emph{maximum a posteriori (MAP) estimation} under a given likelihood \(p(\D | \theta) := \prod_{i=1}^m p(y^{(i)} | x^{(i)}, \theta)\) and a prior \(p(\theta)\).

Let \(\Delta^c\) be the \((c-1)\)-probability simplex.
We define the \emph{softmax} inverse-link function \(\sigma: \R^c \to \Delta^c\) by \(\sigma_k(z) := \exp(z_k)/\sum_{i=1}^c \exp(z_i)\) for each \(k = 1, \dots, c\).
A common choice of the likelihood function for \(c\)-class classification networks is the softmax-Categorical likelihood:
Given a datum pair \((x, y) \in \D\) and parameter \(\theta\), the mapped output \(\sigma(F(x; \theta))\) can be interpreted as a probability vector and the \emph{Categorical log-likelihood} over it can be defined by \(\log p_\text{Cat}(y | x, \theta) := \log \sigma_{y_k}(F(x; \theta))\), i.e.\ it is simply the logarithm of the \(y_k\)-th component of the network's softmax output.
Considering all the data, we write \(\log p_\text{Cat}(\D | \theta) := \sum_{i=1}^m \log p_\text{Cat}(y^{(i)} | x^{(i)}, \theta)\).
Then, given a prior \(p(\theta)\), the MAP estimate is a particular \(\theta\) satisfying \(\argmax_\theta \log p_\text{Cat}(\D | \theta) + \log p(\theta) = \argmax_\theta \log p(\theta | \D)\).
In the case of a zero-mean isotropic Gaussian prior, this corresponds to a maximum likelihood under a weight decay regularization.

While MAP-estimated networks are established to achieve high accuracy, they are overconfident \citep{nguyen2015deep}.
Bayesian methods promise to mitigate this issue \citep[etc.]{mackay1992evidence,wilson2020bayesian}.
The core idea of \emph{Bayesian neural networks (BNNs)} is to infer the full posterior \(p(\theta | \D)\) (instead of just a single estimate of \(\theta\) in the case of MAP estimation) and marginalizing it to obtain the predictive distribution \(p(y | x, \D) = \int \sigma(F(x; \theta)) \, p(\theta | \D) \,d\theta\).
Unfortunately, due to the nonlinearity of \(F\) in \(\theta\), this exact posterior is intractable, and various approximations have been proposed.
Prominent among them are variational Bayes \citep[VB,][etc.]{hinton1993keeping,graves_practical_2011,blundell_weight_2015} and Laplace approximations \citep[LAs,][etc.]{mackay1992practical,ritter_scalable_2018}.
Once an approximation \(q(\theta) \approx p(\theta | \D)\) has been obtained, one can easily use it to approximate \(p(y | x, \D)\) via Monte Carlo (MC) integration.

  \section{WHY BNNS UNDERPERFORM}\label{sec:analysis}

Let \(U\) be the \emph{data region}, i.e.\ a subset of the input space \(\R^n\) where the distribution \(P(X)\) assigns non-negligible mass.
Suppose \(V := \R^n \setminus U\) is the remaining subset of the input space \(\R^n\) that has low mass under \(P(X)\), i.e.\ it is the \emph{OOD region}.
It has recently been shown that any point estimate of \(F\) can induce an arbitrarily overconfident prediction on \(V\) \citep{hein2019relu,nguyen2015deep}.
While Bayesian methods have been shown to ``fix'' this issue in the asymptotic regime \citep[i.e. when the distance between a test point \(x \in V\) and the data region \(U\) tends to infinity; see][]{kristiadi2020being,kristiadi2020infinite}, no such guarantee has been shown for \emph{non-asymptotic} regime which contains outliers that are relatively close to \(U\).
In fact, empirical evidence shows that BNNs often yield suboptimal results in this regime, as \cref{fig:one} shows.

In an adjacent field, the frequentist community has proposed a technique---referred to here as \emph{OOD training}---to address this issue.
The core idea is to ``expose'' the network to a particular kind of OOD data and let it generalize to unseen outliers.
Suppose \(\Dout := \{ \widehat{x}_i \in V \}_{i=1}^{m_\text{out}}\) is a collection of \(m_\text{out}\) points sampled from some distribution on \(V\).
Then, one can incorporate these OOD samples into the standard MAP objective via an additional objective function \(\mathcal{L}\) that depends on the network and \(\Dout\), but not the dataset \(\D\).
We thus do the following:
\begin{equation} \label{eq:ood_obj}
    \argmax_\theta\, \log p_\text{Cat}(\D | \theta) + \log p(\theta) + \mathcal{L}(\theta; \Dout) .
\end{equation}
For instance, \citet{hendrycks2018deep} define \(\mathcal{L}\) to be the negative cross-entropy between the softmax output of \(F\) under \(\Dout\) and the uniform discrete distribution.
Intuitively, \eqref{eq:ood_obj} tries to find a parameter vector of \(F\) that induces well-calibrated predictions both inside and outside of \(U\).
In particular, ideally, the network should retain the performance of the MAP estimate in \(U\), while attaining the maximum entropy or uniform confidence prediction everywhere in \(V\).
Empirically, this frequentist robust training scheme obtains state-of-the-art performance in OOD detection benchmarks \citep[etc.]{hendrycks2018deep,meinke2020towards,bitterwolf2020certifiably}.

While some works have employed OOD data for tuning the hyperparameters of BNNs \citep{kristiadi2020being,kristiadi2020infinite}, ultimately OOD data are not considered as a first-class citizen in the Bayesian inference itself.
Furthermore, while one can argue that theoretically, BNNs can automatically assign high uncertainty over \(V\) and thus robust to outliers, as we have previously discussed, empirical evidence suggests otherwise.
Altogether, it thus now seems likely that indeed the fact that BNNs are not exposed to OOD data during training is a major factor contributing to the discrepancy in their UQ performance compared to that of the state-of-the-art frequentist methods.

  \section{OOD TRAINING FOR BNNS}\label{sec:method}

Motivated by the hypothesis laid out in the previous section, our goal here is to come up with an OOD training scheme for standard BNN inference while retaining a reasonable Bayesian interpretation.
To this end, we explore four different ways of incorporating OOD data in Bayesian inference by making different assumptions about the data and hence the likelihood, starting from the most philosophically clean to the most heuristic.\footnote{One might be tempted to treat \(\mathcal{L}\) in \eqref{eq:ood_obj} as a log-prior. However not only does this mean that we have a (controversial) data-dependent prior, but also it introduces implementation issues, e.g. the KL-divergence term in VB's objective cannot be computed easily anymore.}

\subsection*{Method 1: Extra ``None Class''}

The most straightforward yet philosophically clean way to incorporate unlabeled OOD data is by adding an extra class, corresponding to the ``none class''---also known as the ``dustbin'' or ``garbage class'' \citep{zhang2017universum}.
That is, we redefine our network \(F\) as a function \(\R^n \times \R^{d + \widehat{d}} \to \R^{c+1}\) where \(\widehat{d}\) is the number of additional parameters in the last layer associated with the extra class.
Note that this is different from the ``background class'' method \citep{zhang2017universum,wang2021statistical} which assumes that the extra class is tied to the rest of the classes and thus does not have additional parameters.
We choose to use the ``dustbin class'' since \citet{zhang2017universum} showed that it is the better of the two.

Under this assumption, we only need to label all OOD data in \(\Dout\) with the class \(c+1\) and add them to the true dataset \(\D\).
That is, the new dataset is \(\widetilde{\D} := \D \amalg \{ (\xout^{(1)}, c+1), \dots, (\xout^{(m_\text{out})}, c+1) \}\), where \(\amalg\) denotes disjoint union.
Under this setting, we can directly use the Categorical likelihood, and thus, a BNN with this assumption has a sound Bayesian interpretation.

\subsection*{Method 2: Soft Labels}

In this method, we simply assume that the data have ``soft labels'', i.e.\ the labels are treated as general probability vectors, instead of restricted to integer labels \citep{thiel2008classification}.\footnote{The term ``soft label'' here is different than ``fuzzy label'' \citep{kuncheva2000fuzzy,elgayar2006study} where it is not constrained to sum to one.}
Thus, we can assume that the target \(Y\) is a \(\Delta^c\)-valued random variable.
Under this assumption, since one-hot vectors are also elements of \(\Delta^c\) (they represent the \(c\) corners of \(\Delta^c\)), we do not have to redefine \(\D\) other than to one-hot encode the original integer labels.

Now let us turn our attention to the OOD training data.
The fact that these data should be predicted with maximum entropy suggests that the suitable label for any \(\xout \in \Dout\) is the uniform probability vector \(u := (1/c, \dots, 1/c)\) of length \(c\)---the center of \(\Delta^c\).
Thus, we can redefine \(\Dout\) as the set \(\{ (\xout^{(i)}, u) \}_{i=1}^m\), and then define a new joint dataset \(\widetilde{\D} := \D \amalg \Dout\) containing both the soft-labeled in- and out-distribution training data.
Note that without the assumption that \(Y\) is a simplex-valued random variable, we cannot assign the label \(u\) to the OOD training data, and thus we cannot naturally convey our intuition that we should be maximally uncertain over OOD data.

Under the previous assumption, we have to adapt the likelihood.
A straightforward choice for simplex-valued random variables is the Dirichlet likelihood \(p_\text{Dir}(y | x, \theta) := \text{Dir}(y | \alpha(F(x; \theta)))\) where we have made the dependence of \(\alpha\) to the network output \(F(x; \theta)\) explicit.
So, we obtain the log-likelihood function
\begin{equation} \label{eq:dir_ll}
  \begin{aligned}
    \log p_\text{Dir}(y | x, \theta) = &\log \Gamma \left( \alpha_0 \right) - \log \Gamma(\alpha_k(F(x; \theta))) \\
      &+ \sum_{k=1}^c (\alpha_k(F(x; \theta)) - 1) \log y_k  ,
  \end{aligned}
\end{equation}
where \(\alpha_0 := \sum_{k=1}^c \alpha_k(F(x; \theta))\) and \(\Gamma\) is the Gamma function.
Therefore, the log-likelihood for \(\widetilde{\D}\) is given by \(\log p_\text{Dir}(\widetilde{\D} | x, \theta) = \sum_{i=1}^{m} \log p_\text{Dir}(y^{(i)} | x^{(i)}, \theta) + \sum_{i=1}^{m_\text{out}} \log p_\text{Dir}(u | \xout^{(i)}, \theta)\), which can readily be used in a Bayesian inference.

One thing left to discuss is the definition of \(\alpha(F(x; \theta))\).
An option is to decompose it into the mean and precision \citep{minka2000estimating}.
We do so by writing \(\alpha_k(F(x; \theta)) = \gamma \, \sigma_k(F(x; \theta))\) for each \(k = 1, \dots, c\), where \(\gamma\) is the precision (treated as a hyperparameter) and the softmax output \(\sigma(F(x; \theta))\) represents the mean---which is valid since it is an element of \(\Delta^c\).
The benefits are two-fold: First, since we focus solely on the mean, it is easier for optimization \citep{minka2000estimating}.
Indeed, we found that the alternatives, such as \(\alpha_k(F(x; \theta)) = \exp(F_k(x; \theta))\) yield worse results.
Second, after training, we can use the softmax output of \(F\) as usual without additional steps, i.e.\ when making a prediction, we can treat the network as if it was trained using the standard softmax-Categorical likelihood.

\subsection*{Method 3: Mixed Labels}

There is a technical issue when using the Dirichlet likelihood for the in-distribution data: It is known that the Dirichlet likelihood does not work well with one-hot encoded vectors and that it is harder to optimize than the Categorical likelihood \citep{malinin2018predictive}.
To see this, notice in \eqref{eq:dir_ll} that the logarithm is applied to \(y_k\), in contrast to \(\sigma_k(F(x; \theta))\) in the Categorical likelihood.
If \(y\) is a one-hot encoded vector, this implies that for all but one \(k \in \{ 1, \dots, c \}\), the expression \(\log y_k\) is undefined and thus the entire log-likelihood also is.
While one can mitigate this issue via e.g.\ label smoothing \citep{malinin2018predictive,szegedy2016rethinking}, ultimately we found that models with the Dirichlet likelihood generalize worse than their Categorical counterparts (more in \cref{sec:experiments}).
Fortunately, the Dirichlet log-likelihood \eqref{eq:dir_ll} does not suffer from this issue when used for OOD data because their label \(u\) is the uniform probability vector---in particular, all components of \(u\) are strictly larger than zero.

Motivated by these observations, we combine the best of both worlds in the stability of the Categorical likelihood in modeling ``hard'' one-hot encoded labels (or equivalently, integer labels) and the flexibility of the Dirichlet likelihood in modeling soft labels.
To this end, we assume that all the in-distribution data in \(\D\) have the standard integer labels, while all the OOD data in \(\Dout\) have soft labels.
Then, assuming \(\widetilde{\D} = \D \amalg \Dout\), we define the following ``mixed'' log-likelihood:
\begin{equation*}
  \begin{aligned}
    \log p(\widetilde{\D} | \theta) := &\sum_{i=1}^{m} \log p_\text{Cat}(y^{(i)} | x^{(i)}, \theta) \\
      &+ \sum_{i=1}^{m_\text{out}} \log p_{\text{Dir}}(u | \xout^{(i)}, \theta) .
  \end{aligned}
\end{equation*}
The implicit assumption of this formulation is that, unlike the two previous methods, we have two distinct generative processes for generating the labels of input points in \(U\) and \(V\).
Data in \(\D\) can thus have a different ``data type'' than data in \(\Dout\).
This method can therefore be interpreted as solving a multi-task or multi-modal learning problem.

\subsection*{Method 4: Frequentist-Loss Likelihood}

Considering the effectiveness of frequentist methods, it is thus tempting to give a direct Bayesian treatment upon them.
But to do so, we first have to find a sound probabilistic justification of \(\mathcal{L}\) in \eqref{eq:ood_obj} since not all loss functions can be interpreted as likelihood.
We use the OE objective \citep{hendrycks2018deep} as a use case.

First, recall that OE's OOD objective---the last term in \eqref{eq:ood_obj}---is given by
\begin{equation} \label{eq:oe_loss}
  \begin{aligned}
    \mathcal{L}_\text{OE}(\theta; &\Dout) := - \E_{\xout \sim \Dout} (H(\sigma(F(\xout; \theta)), u)) \\
      &= \frac{1}{c \, m_\text{out}} \sum_{i=1}^{m_\text{out}} \sum_{k=1}^c \log \sigma_c(F(\xout^{(i)}; \theta)) ,
  \end{aligned}
\end{equation}
where \(u := (1/c, \dots, 1/c) \in \Delta^c\) is the uniform probability vector of length \(c\) and \(H\) is the functional measuring the \emph{cross-entropy} between its two arguments.
Our goal here is to interpret \eqref{eq:oe_loss} as a log-likelihood function: we aim at finding a log-likelihood function \(\log p(\Dout | \theta)\) over \(\Dout\) that has the form of \(\mathcal{L}_\text{OE}\). This is sufficient for defining the overall likelihood over \(\D\) and \(\Dout\) since given this function and assuming that these datasets are independent, we readily have a probabilistic interpretation of the log-likelihood terms in \eqref{eq:ood_obj}: \(\log p(\D, \Dout | \theta) = \log p_\text{Cat}(\D | \theta) + \log p(\Dout | \theta)\).

\begin{figure*}
  \begin{minipage}{0.945\textwidth}
    \subfloat{\includegraphics[width=0.2\textwidth]{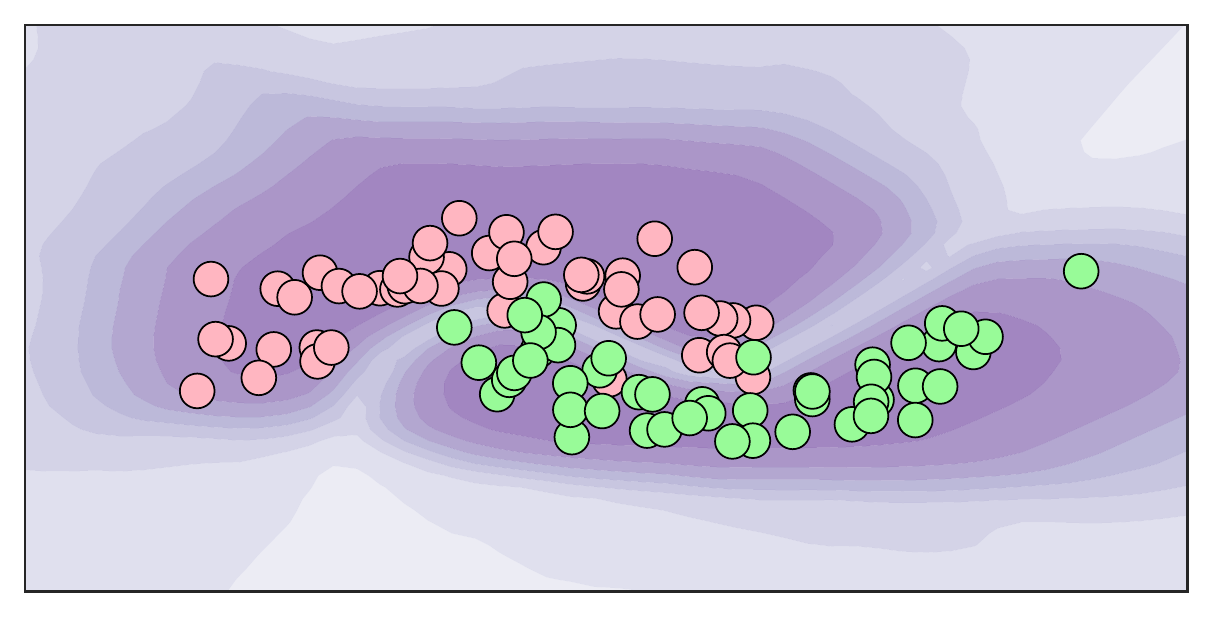}}
    \subfloat{\includegraphics[width=0.2\textwidth]{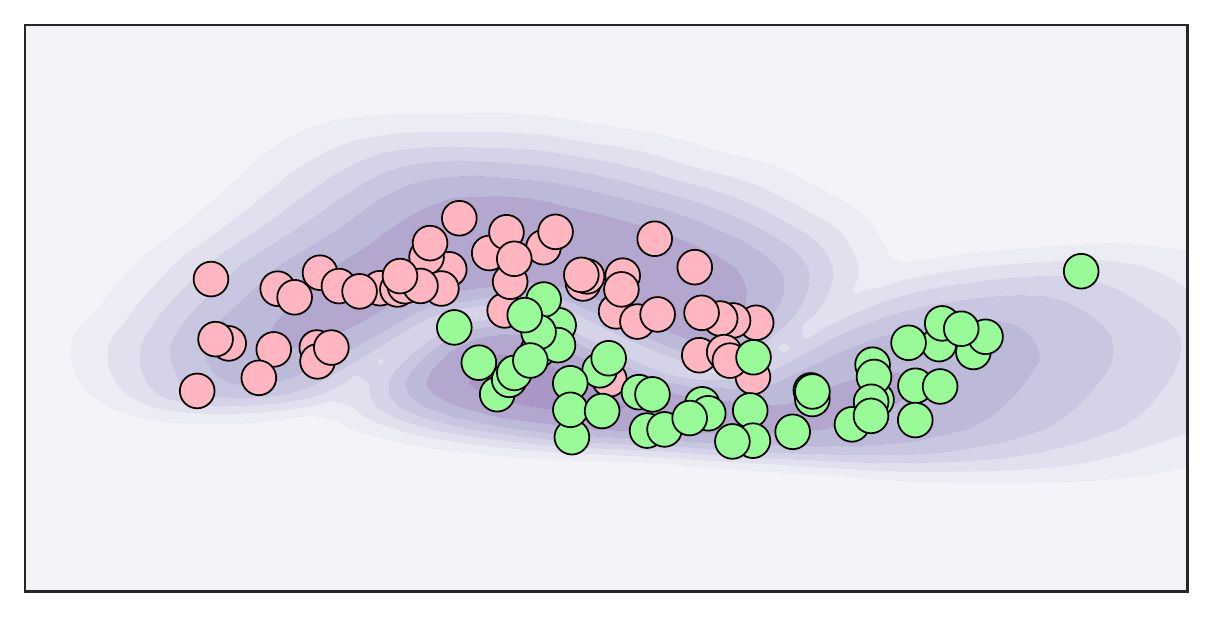}}
    \subfloat{\includegraphics[width=0.2\textwidth]{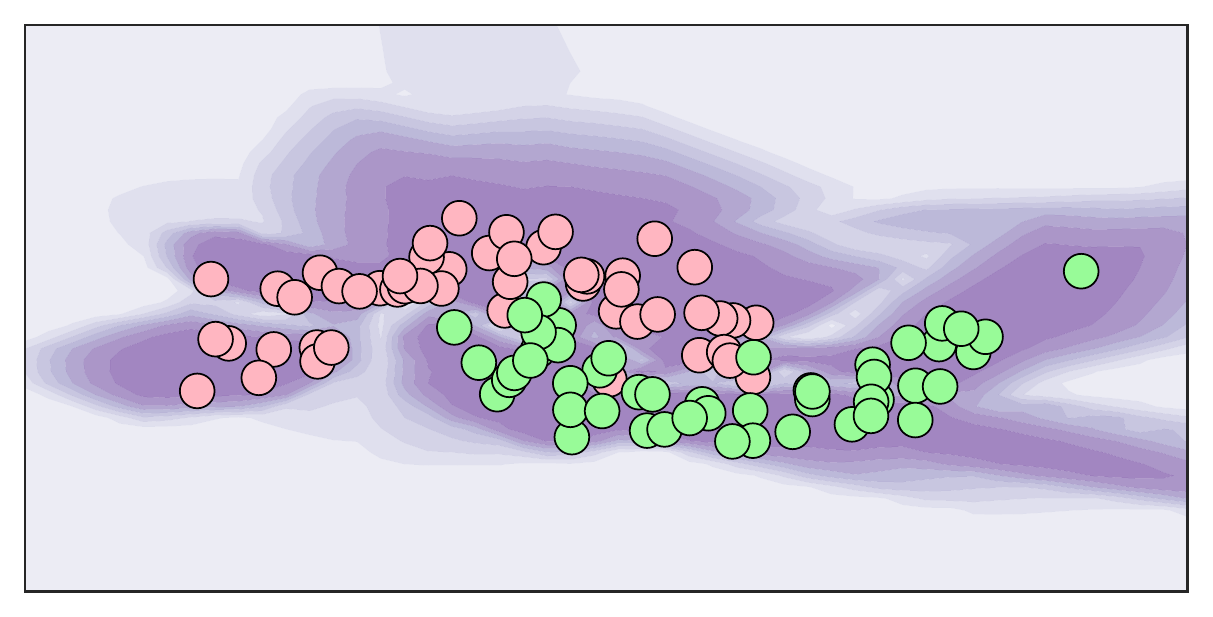}}
    \subfloat{\includegraphics[width=0.2\textwidth]{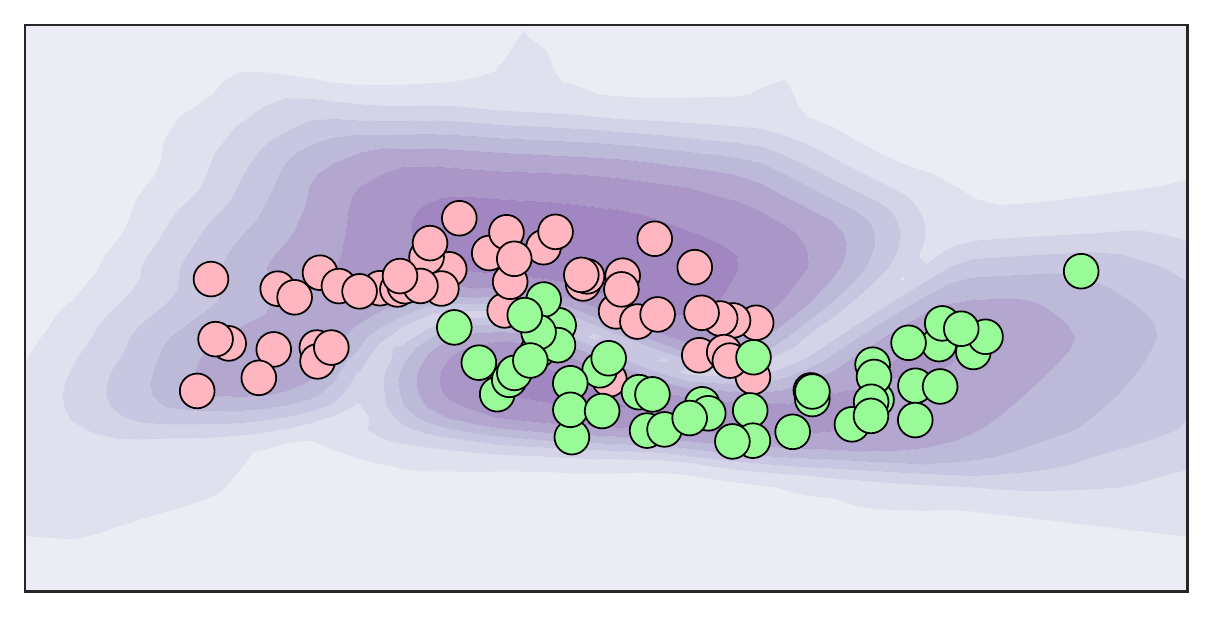}}
    \subfloat{\includegraphics[width=0.2\textwidth]{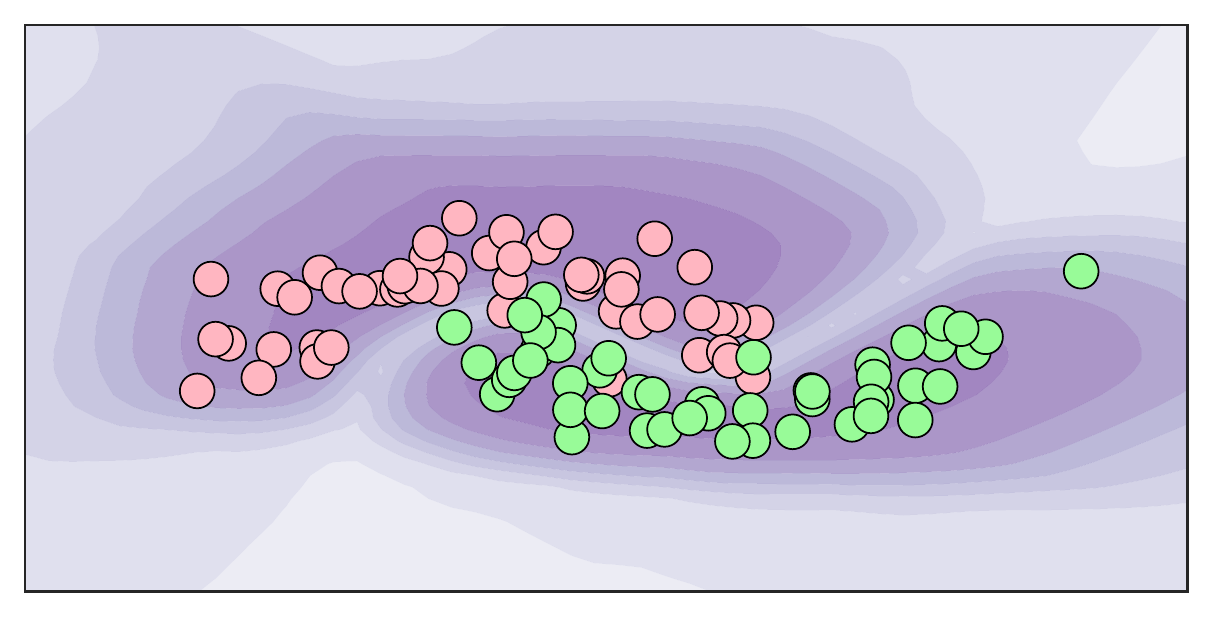}}

    \vspace{-1em}
    \setcounter{subfigure}{0}

    \subfloat[Freq. OE]{\includegraphics[width=0.2\textwidth]{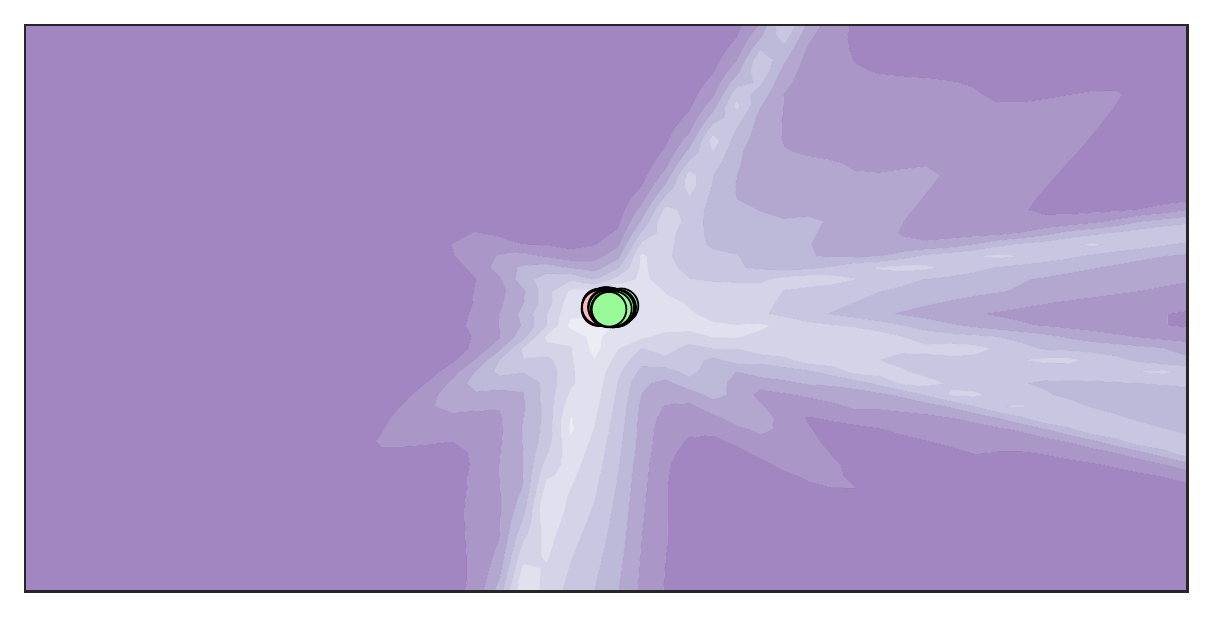}}
    \subfloat[LA+NC]{\includegraphics[width=0.2\textwidth]{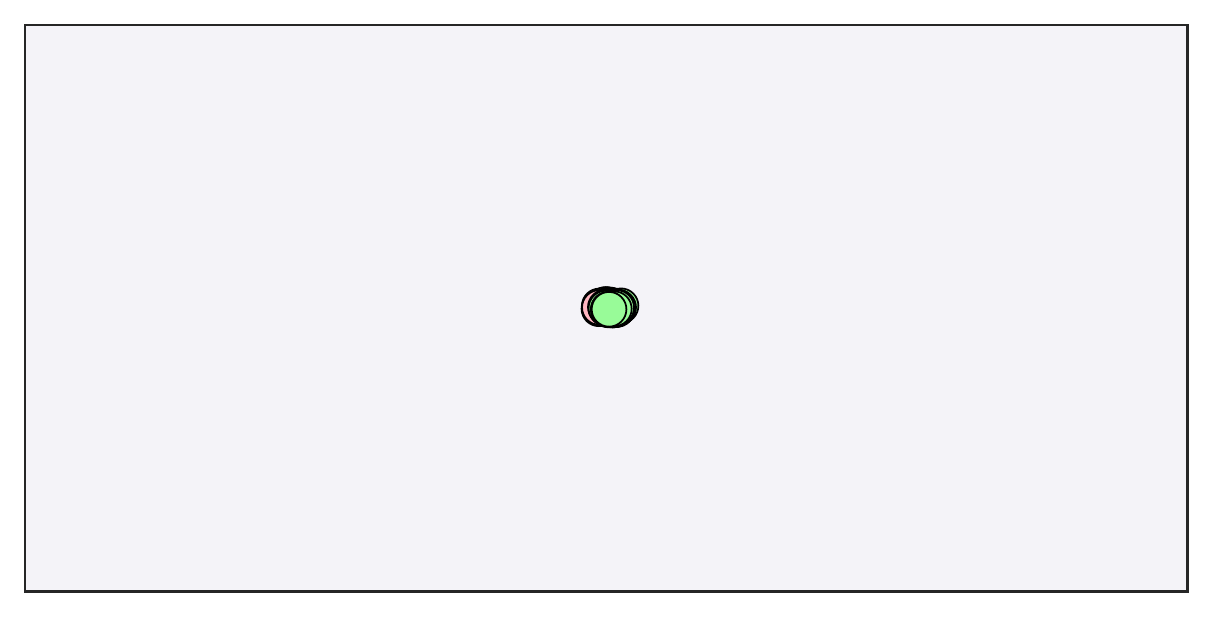}}
    \subfloat[LA+SL]{\includegraphics[width=0.2\textwidth]{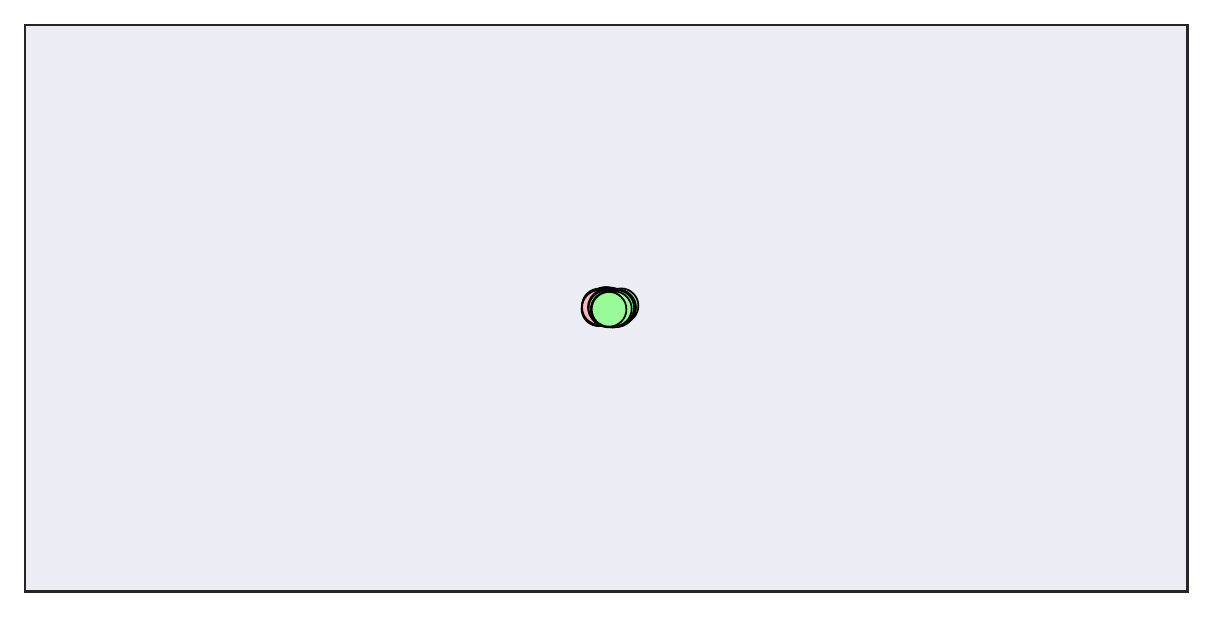}}
    \subfloat[LA+ML]{\includegraphics[width=0.2\textwidth]{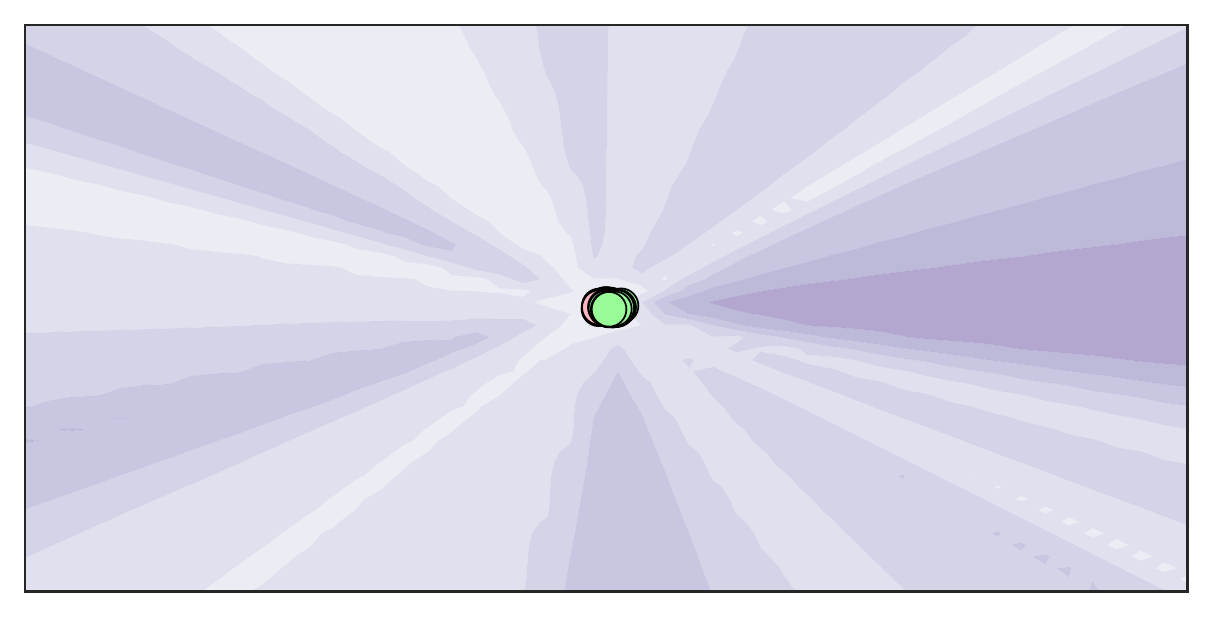}}
    \subfloat[LA+OE]{\includegraphics[width=0.2\textwidth]{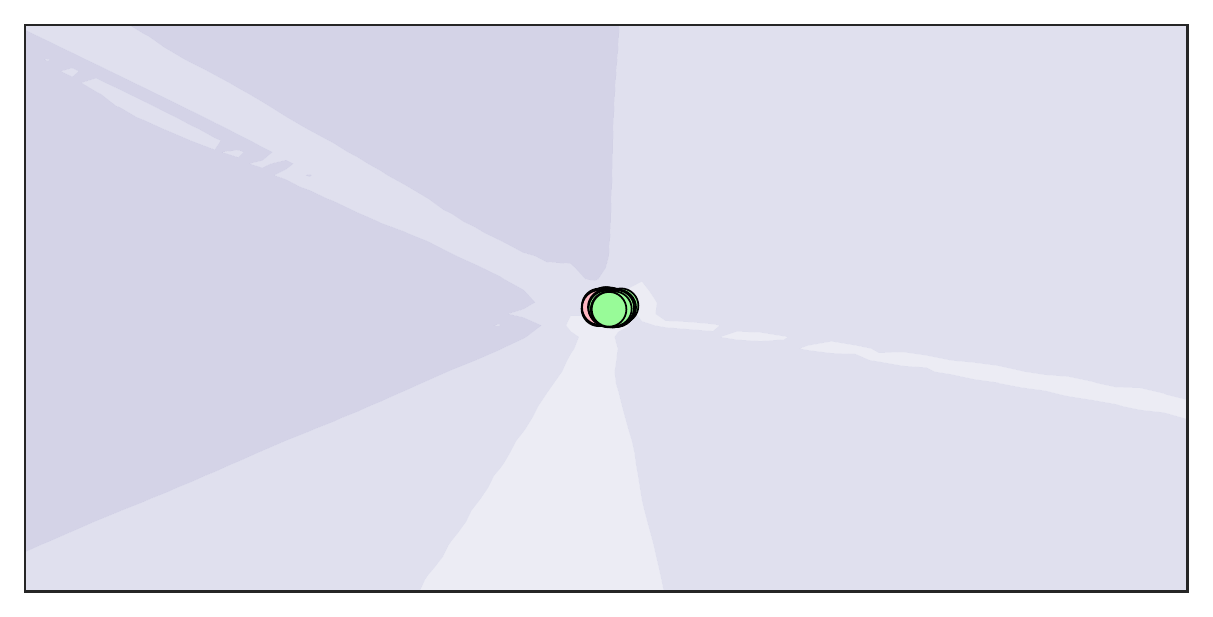}}
  \end{minipage}
  \hspace{-1em}
  \begin{minipage}{0.04\textwidth}
    \vspace{-0.29em}
    \subfloat{\includegraphics{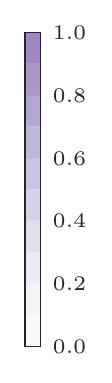}}
  \end{minipage}

  \caption{
    Confidence estimate of the frequentist OE baseline \textbf{(a)} and the Bayesian OOD-training methods discussed in \cref{sec:method} \textbf{(b-e)}.
    For the BNNs, we use the Laplace approximation (LA).
    The suffixes ``+NC'', ``+SL'', ``+ML'', and ``+OE'' refer to the ``none class'', ``soft labels'', ``mixed labels'', and ``OE likelihoods'' methods, respectively.
    The bottom row shows the zoomed-out versions of the top one.
    The OOD training data is sampled from \(\mathrm{Unif}[-6, 6]^2\).
    While OE yields calibrated uncertainty near the training data (a), it is overconfident far from them.
    Meanwhile, Bayesian methods are less overconfident far away from the data (b-e).}
  \label{fig:toy}
\end{figure*}

We begin with the assumption that the Categorical likelihood is used to model both the in- and out-of-distribution data---in particular, we use the standard integer labels for both \(\D\) and \(\Dout\).
Now, recall that the OOD data ideally have the uniform confidence, that is, they are equally likely under all possible labels.
But since we have assumed hard labels, we cannot use \(u\) directly as the label for \(\xout \in \Dout\).
To circumvent this, we redefine the OOD dataset \(\Dout\) by assigning all \(c\) possible labels to each \(\xout\), based on the intuition that \(F\) should be ``maximally confused'' about the correct label of \(\xout\).
\begin{equation} \label{eq:oe_data_replication}
  \begin{aligned}
    \Dout := \{ &(x_\text{out}^{(1)}, 1), \dots, (x_\text{out}^{(1)}, K), \dots, \\
      &(x_\text{out}^{(m_\text{out})}, 1), \dots, (x_\text{out}^{(m_\text{out})}, K) \} .
  \end{aligned}
\end{equation}
Thus, given \(m_\text{out}\) unlabeled OOD data, we have \(\abs{\Dout} = c \, m_\text{out}\) OOD data points in our OOD training set.
So, the negative log-Categorical likelihood over \(\Dout\) is given by
\begin{equation}
    -\log p(\Dout | \theta) = \sum_{i=1}^{m_\text{out}} \sum_{k=1}^c \log \sigma_k(F(\xout^{(i)}; \theta))
\end{equation}
Comparing this to \eqref{eq:oe_loss}, we identify that \(\log p(\Dout | \theta)\) is exactly \(\mathcal{L}_\text{OE}\), up to a constant factor \(1/(c \, m_\text{out})\), which can be thought of as a tempering factor to \(p(\Dout | \theta)\).
We have thus obtained the probabilistic interpretation of OE's objective---this likelihood can then be soundly used in a Bayesian inference---albeit arising from applying a heuristic \eqref{eq:oe_data_replication} to the data.

\subsection{Remark}

Here, we consider the question of whether there is an inherent advantage of using OOD-trained BNNs instead of the standard OOD-trained network.
One answer to this question is given by the recent finding that Bayesian methods naturally yield low uncertainty in regions far away from the data \citep{kristiadi2020being}.
In contrast, OOD-trained point-estimated networks do not enjoy such a guarantee by default and must resort to e.g.\ generative models \citep{meinke2020towards}.
We illustrate this observation synthetically in \cref{fig:toy}.

  \section{RELATED WORK}\label{sec:related}

OOD training for BNNs has recently been used for tuning the hyperparameters of LAs \citep{kristiadi2020being,kristiadi2020learnable}.
However, it appears that OOD training is not commonly utilized by BNNs in the Bayesian inference itself.
\citet{wang2021statistical}, at the same time window as this work, also proposed OOD training for BNNs by justifying the presence of OOD training data as a consequence of the data curation process.
Nevertheless, their method is different than all the methods proposed here, as discussed in \cref{sec:method}, and only validated on a single BNN.
Meanwhile, we explore four distinct methods and extensively validate them on various BNNs, see \cref{sec:experiments}.

From an adjacent field, adversarial training for BNNs has recently been studied.
In particular, \citet{liu2018advbnn} specifically employ VB and modify the first term of the ELBO to take into account the worst-case perturbation of each data point, which can be thought of as a particular type of OOD data.
Unlike theirs, our methods are for general OOD data and are agnostic to the approximate inference method.

Non-Bayesian Dirichlet-based models have recently been studied for UQ \citep{sensoy2018evidential}.
Similar to our proposed ``soft labels'' and ``mixed labels'' likelihoods, \citet{malinin2018predictive,malinin2019reverse,nandy2020towards} use the Dirichlet distribution as the output of a non-Bayesian network and employ OOD training via a custom, non-standard loss.
Their methods' Bayesian interpretation is therefore unclear.
In contrast, for modeling soft labels, we simply use the standard Dirichlet log-likelihood function, which is well-studied in the context of generalized linear models \citep{gueorguieva2008dirichlet}. Our methods thus retain a clear Bayesian interpretation when used in BNNs.

  \section{EXPERIMENTS}\label{sec:experiments}

\begin{table*}[t]
  \caption{Test accuracy (\(\uparrow\)) / ECE (\(\downarrow\)), averaged over five prediction runs. Best values in each categories (separated by horizontal lines) are in bold.}
  \label{tab:acc_ece}

  \centering
  \small
  \renewcommand{\tabcolsep}{10pt}

  \resizebox{\textwidth}{!}{
    \begin{tabular}{lllllll}
      \toprule
      \textbf{\footnotesize} & {\footnotesize\bf MNIST} & {\footnotesize\bf F-MNIST} & {\footnotesize\bf SVHN} & {\footnotesize\bf CIFAR-10} & {\footnotesize\bf CIFAR-100} \\
      \midrule

      MAP & 99.4$\pm$0.0\,/\,6.4$\pm$0.0 & 92.4$\pm$0.0\,/\,13.9$\pm$0.0 & 97.4$\pm$0.0\,/\,8.9$\pm$0.0 & 94.8$\pm$0.0\,/\,10.0$\pm$0.0 & 76.7$\pm$0.0\,/\,14.3$\pm$0.0 \\
      DE & \textbf{99.5}$\pm$0.0\,/\,8.6$\pm$0.0 & \textbf{93.6}$\pm$0.0\,/\,3.6$\pm$0.0 & \textbf{97.6}$\pm$0.0\,/\,\textbf{3.5}$\pm$0.0 & \textbf{95.7}$\pm$0.0\,/\,\textbf{4.5}$\pm$0.0 & \textbf{80.0}$\pm$0.0\,/\,\textbf{1.9}$\pm$0.0 \\
      OE & 99.4$\pm$0.0\,/\,\textbf{5.3}$\pm$0.0 & 92.3$\pm$0.0\,/\,12.1$\pm$0.0 & 97.4$\pm$0.0\,/\,10.6$\pm$0.0 & 94.6$\pm$0.0\,/\,13.2$\pm$0.0 & 76.7$\pm$0.0\,/\,15.0$\pm$0.0 \\

      \midrule

      VB & \textbf{99.5}$\pm$0.0\,/\,11.2$\pm$0.3 & 92.4$\pm$0.0\,/\,3.7$\pm$0.2 & 97.5$\pm$0.0\,/\,5.7$\pm$0.2 & 94.9$\pm$0.0\,/\,5.8$\pm$0.2 & \textbf{75.4}$\pm$0.0\,/\,\textbf{8.3}$\pm$0.0 \\
      +NC & 99.4$\pm$0.0\,/\,12.6$\pm$0.3 & 92.2$\pm$0.0\,/\,3.3$\pm$0.1 & 97.5$\pm$0.0\,/\,\textbf{4.1}$\pm$0.1 & 94.4$\pm$0.0\,/\,5.5$\pm$0.1 & 74.1$\pm$0.0\,/\,10.7$\pm$0.1 \\
      +SL & \textbf{99.5}$\pm$0.0\,/\,10.5$\pm$0.3 & \textbf{93.1}$\pm$0.0\,/\,6.3$\pm$0.1 & \textbf{97.6}$\pm$0.0\,/\,9.3$\pm$0.2 & 93.0$\pm$0.0\,/\,11.0$\pm$0.1 & 71.4$\pm$0.0\,/\,13.0$\pm$0.0 \\
      +ML & 99.3$\pm$0.0\,/\,11.8$\pm$0.2 & 92.0$\pm$0.0\,/\,\textbf{2.5}$\pm$0.1 & \textbf{97.6}$\pm$0.0\,/\,4.2$\pm$0.0 & \textbf{95.0}$\pm$0.0\,/\,4.9$\pm$0.2 & \textbf{75.4}$\pm$0.0\,/\,10.4$\pm$0.0 \\
      +OE & 99.4$\pm$0.0\,/\,\textbf{10.0}$\pm$0.2 & 92.3$\pm$0.0\,/\,3.0$\pm$0.2 & \textbf{97.6}$\pm$0.0\,/\,5.7$\pm$0.2 & 94.8$\pm$0.0\,/\,\textbf{4.6}$\pm$0.2 & 74.2$\pm$0.0\,/\,8.9$\pm$0.0 \\

      \midrule

      LA & 99.4$\pm$0.0\,/\,7.6$\pm$0.1 & 92.5$\pm$0.0\,/\,11.3$\pm$0.2 & 97.4$\pm$0.0\,/\,3.3$\pm$0.3 & \textbf{94.8}$\pm$0.0\,/\,7.5$\pm$0.3 & 76.6$\pm$0.1\,/\,8.3$\pm$0.1 \\
      +NC & 99.4$\pm$0.0\,/\,5.4$\pm$0.7 & 92.4$\pm$0.0\,/\,8.5$\pm$0.3 & 97.3$\pm$0.0\,/\,4.6$\pm$0.2 & 94.0$\pm$0.0\,/\,\textbf{6.6}$\pm$0.3 & 76.2$\pm$0.0\,/\,6.1$\pm$0.0 \\
      +SL & \textbf{99.7}$\pm$0.0\,/\,12.1$\pm$1.1 & \textbf{93.2}$\pm$0.0\,/\,\textbf{3.2}$\pm$0.3 & \textbf{97.5}$\pm$0.0\,/\,7.4$\pm$0.2 & 93.6$\pm$0.0\,/\,10.2$\pm$0.2 & 72.3$\pm$0.1\,/\,7.1$\pm$0.2 \\
      +ML & 99.4$\pm$0.0\,/\,7.5$\pm$1.0 & 92.5$\pm$0.0\,/\,5.9$\pm$0.2 & 97.4$\pm$0.0\,/\,\textbf{2.9}$\pm$0.2 & \textbf{94.8}$\pm$0.0\,/\,6.9$\pm$0.3 & 76.5$\pm$0.1\,/\,\textbf{4.4}$\pm$0.1 \\
      +OE & 99.4$\pm$0.0\,/\,\textbf{4.8}$\pm$0.7 & 92.3$\pm$0.0\,/\,7.4$\pm$0.1 & 97.4$\pm$0.0\,/\,3.2$\pm$0.1 & 94.6$\pm$0.0\,/\,8.8$\pm$0.1 & \textbf{76.7}$\pm$0.1\,/\,\textbf{4.4}$\pm$0.1 \\

      \bottomrule
    \end{tabular}
  }
\end{table*}

\subsection{Setup}

\paragraph*{Baselines}
We use the following strong, recent baselines to represent non-Bayesian methods:\footnote{Deep Ensemble can also be seen as a Bayesian method, but it was originally proposed as a frequentist method.} (i) standard MAP-trained network (\textbf{MAP}), (ii) Deep Ensemble \citep[\textbf{DE},][]{lakshminarayanan2017simple}, and (iii) Outlier Exposure \citep[\textbf{OE},][]{hendrycks2018deep}.
Note that DE and OE are among the established state-of-the-art frequentist UQ methods.

For standard Bayesian methods, i.e.\ those considering only \(\D\) in the inference, we use (iv) the all-layer diagonal Laplace approximation on top of the MAP network (\textbf{LA}) and (v) the last-layer mean-field VB \citep[\textbf{VB},][]{graves_practical_2011,blundell_weight_2015}.
We mainly use only these simple Bayesian methods to validate that the proposed likelihood could make even these crudely approximated BNNs competitive to the strong baselines. Results with more advanced BNNs are in Tab.~\ref{tab:ood_aux}.

To represent our methods, we again use the same LA and VB but with the modifications proposed in \cref{sec:method}.
We denote these modified methods \textbf{LA+X} and \textbf{VB+X}, respectively.
Here, \textbf{X} is the abbreviation for the proposed methods, i.e. \textbf{NC} for ``none class'' (Method 1), \textbf{SL} for ``soft label'' (Method 2), \textbf{ML} for ``mixed label'' (Method 3), and \textbf{OE} for ``OE likelihood'' (Method 4).

Finally, we use the LeNet and WideResNet-16-4 architectures, trained in the usual manner---see \cref{sec:app:training_details}.
Source code is available at \url{https://github.com/wiseodd/bayesian_ood_training}.

\vspace{-0.5em}
\paragraph*{Datasets}
As the in-distribution datasets, we use: (i) MNIST, (ii) Fashion-MNIST (F-MNIST), (iii) SVHN, (iv) CIFAR-10, and (v) CIFAR-100.
For each of them, we obtain a validation set of size 2000 by randomly splitting the test set.
For methods requiring OOD training data, i.e.\ OE, LA+X, and VB+X, we use the 32\(\times\)32 downsampled ImageNet dataset \citep{chrabaszcz2017downsampled} as an alternative to the 80M Tiny Images dataset used by \citet{hendrycks2018deep,meinke2020towards}, since the latter is not available anymore.\footnote{\url{https://groups.csail.mit.edu/vision/TinyImages/}.}
For OOD detection tasks, we use various \emph{unseen} (i.e.\ not used for training or tuning) OOD test sets as used in \citep{meinke2020towards,hein2019relu}, both real-world (e.g. E-MNIST) and synthetic (e.g.\ uniform noise).
For text classification, we use the Stanford Sentiment Treebank \citep[SST,][]{socher2013recursive} and the TREC dataset \citep{voorhees2001overview}. We detail of all OOD test sets in \cref{sec:app:ood_dataset}.
We furthermore test the methods in a dataset-shift robustness task using the corrupted CIFAR-10 (CIFAR-10-C) dataset \citep{ovadia2019can,hendrycks2019benchmarking}.

\vspace{-0.5em}
\paragraph*{Metrics}
To measure OOD detection performance, we use the standard FPR95 metric, which measures the false positive rate at 95\% true positive rate.
Other metrics such as average confidence and area under the ROC curve are presented in the appendix.
Meanwhile, to measure dataset-shift robustness and predictive performance, we use test accuracy and expected calibration error (ECE) with 15 bins \citep{naeini2015obtaining}.

\begin{table*}[t]
  \caption{
    OOD data detection in terms of FPR95.
    Values are averages over six OOD test sets and five prediction runs---lower is better. The best values of each category are in bold. Details are in \cref{tab:fpr95} in the appendix.
  }
  \label{tab:ood_summary}

  \centering
  \small
  \renewcommand{\tabcolsep}{10pt}

  \begin{tabular}{lrrrrrr}
    \toprule

    \textbf{Methods} & & {\bf MNIST} & {\bf F-MNIST} & {\bf SVHN} & {\bf CIFAR-10} & {\bf CIFAR-100} \\
    \midrule

    MAP & & 17.7 & 69.4 & 22.4 & 52.4 & 81.0 \\
    DE & & 10.6 & 61.4 & 10.1 & 32.3 & 73.3 \\
    OE & & \textbf{5.4} & \textbf{16.2} & \textbf{2.1} & \textbf{22.8} & \textbf{54.0} \\

    \midrule

    VB & & 25.7 & 63.3 & 22.0 & 36.5 & 77.6 \\
    +NC & & 7.5 & 15.0 & \textbf{1.4} & \textbf{28.0} & \textbf{49.9} \\
    +SL & & \textbf{2.7} & \textbf{4.2} & 1.8 & 40.4 & 62.3 \\
    +ML & & 7.4 & 19.6 & \textbf{1.4} & 29.1 & 50.2 \\
    +OE & & 6.8 & 22.4 & 1.5 & 29.8 & 53.3 \\

    \midrule

    LA & & 19.4 & 68.7 & 17.1 & 53.6 & 81.3 \\
    +NC & & 6.6 & 8.3 & 1.5 & \textbf{20.1} & \textbf{47.4} \\
    +SL & & \textbf{2.2} & \textbf{4.1} & \textbf{1.0} & 38.5 & 60.9 \\
    +ML & & 5.5 & 14.3 & 1.1 & 21.8 & 52.5 \\
    +OE & & 5.4 & 17.0 & 1.1 & 23.3 & 53.9 \\

    \bottomrule
  \end{tabular}
\end{table*}

\subsection{Generalization and Calibration}

We present the generalization and calibration performance in \cref{tab:acc_ece}.
We note that generally, all methods discussed in \cref{sec:method} attain comparable accuracy to, and are better calibrated than the vanilla MAP/OE models.
However, the ``soft label'' method tends to underperform in both accuracy and ECE---this can be seen clearly on CIFAR-100.
This issue appears to be because of the numerical issue we have discussed in \cref{sec:method}.
Note that this issue seems to also plague other Dirichlet-based methods \citep{malinin2018predictive,malinin2019reverse}.
Overall, it appears that Bayesian OOD training with NC, ML, and OE is not harmful to the in-distribution performance---they are even more calibrated than the frequentist OE.

\subsection{OOD Detection}

\begin{table}[t]
  \caption{
    OOD data detection on text classification tasks.
    Values are averages over five prediction runs and additionally, three OOD test sets for FPR95.
    Details are in the appendix (\cref{tab:acc_ece_nlp,tab:fpr95_nlp_detailed}).
  }
  \label{tab:ood_nlp}

  \centering
  \small

  \begin{tabular}{lrrrrrr}
    \toprule

    & & \multicolumn{2}{c}{\bf ECE} & & \multicolumn{2}{c}{\bf FPR95} \\
    \cmidrule(r){3-4} \cmidrule(l){6-7}

    \textbf{Methods} & & {\bf SST} & {\bf TREC} & & {\bf SST} & {\bf TREC} \\

    \midrule

    MAP & & 20.8 & 17.2 & & 100.0 & 96.3 \\
    DE & & \textbf{{2.5}} & 10.6 & & 100.0 & 24.2 \\
    OE & & 13.0 & \textbf{9.4} & & \textbf{{0.0}} & \textbf{{0.0}} \\

    \midrule

    LA & & 21.0 & 17.3 & & 100.0 & 96.4 \\
    +NC & & 17.9 & 18.6 & & \textbf{{0.0}} & \textbf{{0.0}} \\
    +SL & & 17.5 & 10.4 & & 95.3 & 0.8 \\
    +ML & & \textbf{11.4} & 11.5 & & 84.6 & \textbf{{0.0}} \\
    +OE & & 12.8 & \textbf{{8.4}} & & \textbf{{0.0}} & \textbf{{0.0}} \\

    \bottomrule
  \end{tabular}
\end{table}

We present the OOD detection results on image classification datasets in \cref{tab:ood_summary}.\footnote{Refer to \cref{sec:app:add_results} for the detailed, non-averaged results for \cref{tab:ood_summary,tab:ood_nlp,tab:ood_noise,tab:ood_aux}, along with additional metrics.}
As indicated in \cref{fig:one}, OE is in general significantly better than even DE while retaining the computational efficiency of MAP.
The vanilla Bayesian baselines, represented by VB and LA, achieve worse results than DE (and thus OE).
But, when OOD training is employed to train these BNNs using the four methods we considered in \cref{sec:method}, their performance improves.
We observe that all Bayesian OOD training methods generally yield better results than DE and become competitive to OE.
In particular, while the ``soft label'' method (SL) is best for ``easy'' datasets (MNIST, F-MNIST), we found that the simplest ``none class'' method (NC) achieves the best results in general.

In \cref{tab:ood_nlp}, we additionally show the results on text classification datasets.
We found that the OOD training methods consistently improve both the calibration and OOD-detection performance of the vanilla Bayesian methods, making them on par with OE.
As before, the ``none class'' method performs well in OOD detection.
This is a reassuring result since NC is also the most philosophically clean (i.e. requires fewer heuristics) than the other three methods considered.

\begin{table}[t]
  \caption{Average ECE and FPR95 under models trained with synthetic noises as \(\Dout\). Both values are averaged over five prediction runs, and additionally six OOD test sets for FPR95. See \cref{tab:acc_ece_smooth,tab:fpr95_noise_more} in the appendix for details.}
  \label{tab:ood_noise}

  \centering
  \small

  \begin{tabular}{lrrrr}
    \toprule

    \textbf{Methods} & & {\bf SVHN} & {\bf CIFAR-10} & {\bf CIFAR-100} \\

    \midrule
    \midrule

    \rowcolor{tablegray}
    \textbf{ECE \(\downarrow\)} & & & & \\

    MAP & & \textbf{8.9} & \textbf{10.0} & \textbf{14.3} \\
    OE & & \textbf{8.9} & 11.5 & 16.1 \\

    \midrule

    LA & & \textbf{3.3} & 7.5 & 8.3 \\
    +NC & & 5.0 & 8.3 & \textbf{3.8} \\
    +SL & & 13.5 & 16.0 & 4.0 \\
    +ML & & 7.4 & \textbf{7.2} & 3.3 \\
    +OE & & 3.8 & \textbf{7.2} & 8.6 \\

    \midrule
    \midrule

    \rowcolor{tablegray}
    \textbf{FPR95 \(\downarrow\)} & & & & \\

    MAP & & 22.4 & 52.4 & 81.0 \\
    OE & & \textbf{11.4} & \textbf{31.0} & \textbf{60.1} \\

    \midrule

    LA & & 17.1 & 53.6 & 81.3 \\
    +NC & & 10.5 & \textbf{26.4} & 64.5 \\
    +SL & & 93.7 & 37.9 & 68.6 \\
    +ML & & 14.4 & 28.4 & 61.0 \\
    +OE & & \textbf{10.1} & 35.3 & \textbf{56.4} \\

    \bottomrule
  \end{tabular}
\end{table}

\begin{table}[t]
  \caption{Average ECE and FPR95 with more sophisticated base models. ``C-10'' stands for CIFAR-10, while ``C-100'' for CIFAR-100}
  \label{tab:ood_aux}

  \centering
  \small

  \begin{tabular}{lrrrrrr}
    \toprule

    & & \multicolumn{2}{c}{\bf ECE} & & \multicolumn{2}{c}{\bf FPR95} \\
    \cmidrule(r){3-4} \cmidrule(l){6-7}

    \textbf{Methods} & & {\bf C-10} & {\bf C-100} & & {\bf C-10} & {\bf C-100} \\

    \midrule

    Flipout & & 10.9 & 19.8 & & 65.0 & 85.4 \\
    +NC & & \textbf{8.2} & \textbf{13.8} & & \textbf{40.9} & \textbf{56.2} \\

    \midrule

    CSGHMC & & \textbf{1.7} & 4.0 & & 60.3 & 81.0 \\
    +NC & & 6.2 & \textbf{2.4} & & \textbf{25.0} & \textbf{43.0} \\

    \midrule

    DE & & \textbf{4.5} & 1.9 & & 32.3 & 73.3 \\
    +NC & & 4.8 & \textbf{1.7} & & \textbf{17.0} & \textbf{44.4} \\

    \bottomrule
  \end{tabular}
\end{table}

\begin{figure*}[htp!]
  \centering

  \subfloat{\input{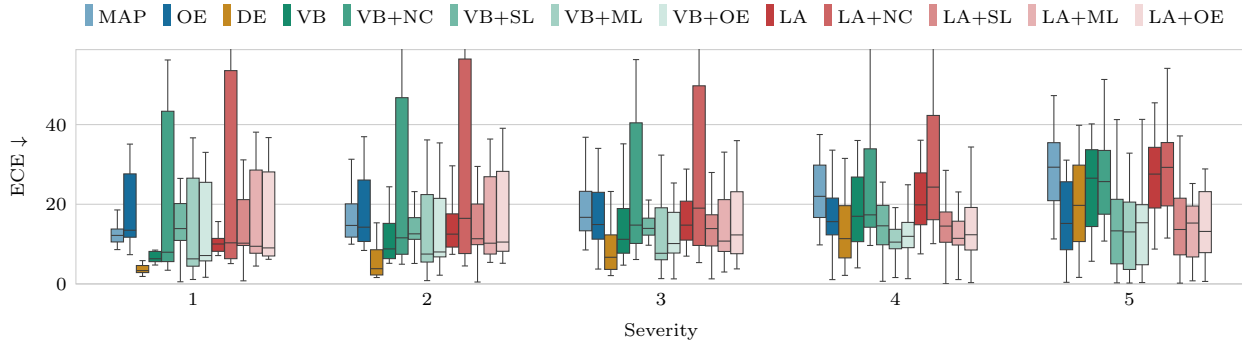}}

  \caption{
    Dataset shift performance on the corrupted-CIFAR-10-C dataset, following \citet{ovadia2019can}. Values are ECE---lower is better.
  }
  \label{fig:dataset_shift}
\end{figure*}

A common concern regarding OOD training is the choice of \(\Dout\).
As an attempt to address this, in \cref{tab:ood_noise} we provide results on OOD detection when the model is trained using a synthetic noise dataset.
The noise dataset used here is the ``smooth noise'' dataset \citep{hein2019relu}, obtained by permuting, blurring, and contrast-rescaling the original training dataset.
We found that even with such a simple OOD dataset, we can still generally obtain better OOD detection results than OE, as shown by the FPR95 values.
Moreover, the combination of Bayesian formalism and OOD training is beneficial in calibrating the in-distribution uncertainty:
We found that using this \(\Dout\) to train OE yields worse-calibrated results than even the vanilla MAP model, as the ECE values show.
In contrast, OOD-trained LA yields better ECE results in general.

Finally, we show that OOD training is beneficial for other BNNs.
In \cref{tab:ood_aux}, we consider two recent (all-layer) BNNs: a VB with the flipout estimator \citep[Flipout,][]{wen2018flipout} and the cyclical stochastic-gradient Hamiltonian Monte Carlo \citep[CSGHMC,][]{zhang2020cyclical}.
Evidently, OOD training improves their OOD detection performance by a large margin.
Moreover, OOD training also improves the performance of DE.

\subsection{Dataset-Shift Robustness}
\label{subsec:ds_robustness}

In this UQ task, OOD training is beneficial for both MAP and the vanilla Bayesian methods (VB, LA), making them competitive to the state-of-the-art DE's performance in larger severity levels, see \cref{fig:dataset_shift}.
Moreover, the OOD-trained VB and LA are in general more calibrated than OE, which shows the benefit of the Bayesian formalism vis-\`{a}-vis the point-estimated OE.
This indicates that \emph{both} being Bayesian and considering OOD data during training are beneficial.

Even though it is the best in OOD detection, here we observe that NC is less calibrated in terms of ECE than its counterparts.
This might be due to the incompatibility of calibration metrics with the additional class: When the data are corrupted, they become closer to the OOD data, and thus NC tends to assign higher probability mass to the last class which does not correspond to any of the true classes (contrast this to other the approaches).
Therefore, in this case, the confidence over the true class becomes necessarily lower---more so than the other approaches.
Considering that calibration metrics depend on the confidence of the true class, the calibration of NC thus suffers.
One way to overcome this issue is to make calibration metrics aware of the ``none class'', e.g. by measuring calibration only on data that have low ``none class'' probability.
We leave the investigation for future research.

\subsection{Costs}

The additional costs associated with all the OOD-training methods presented here are negligible: Like other non-Bayesian OOD-training methods, the only overhead is the additional minibatch of OOD training data at each training iteration---the costs are similar to when considering a standard training procedure with double the minibatch size.
Additionally for LA, in its Hessian computation, one effectively computes it with twice the number of the original data.
However, this only needs to be done \emph{once} post-training.

\subsection{Limitations}

Our methods require a choice of OOD training dataset and it is unclear how can one obtain an OOD training set for specialized problems like medical analyses.
Nevertheless, for image classification or natural language processing, we know that large-scale natural image/text OOD training sets are most useful \citep{hendrycks2018deep}.
Moreover, since our methods are agnostic to the choice of OOD training data, any future advances in the choice of OOD data for OOD training from both the Bayesian and the frequentist communities can be applied to ours.

  \section{CONCLUSION}\label{sec:conclusion}

We raised an important observation regarding contemporary BNNs' performance in uncertainty quantification, in particular in OOD detection tasks:
BNNs tend to underperform compared to non-Bayesian UQ methods.
We hypothesized that this issue is because recent frequentist UQ methods utilize an auxiliary OOD training set.
To validate this hypothesis, we explored ways to incorporate OOD training data into BNNs while still maintaining a reasonable Bayesian interpretation.
Our experimental results showed that using OOD data in approximate Bayesian inference significantly improved the performance of BNNs, making them competitive or even better than their non-Bayesian counterparts.
In particular, we found that the most philosophically Bayesian-compatible way of OOD training---simply add an additional ``none class''---performs best.
We hope that the studied methods can be strong baselines for future work in the Bayesian deep learning community.


  \subsubsection*{Acknowledgments}
  The authors gratefully acknowledge financial support by the European Research Council through ERC StG Action 757275 / PANAMA; the DFG Cluster of Excellence ``Machine Learning - New Perspectives for Science'', EXC 2064/1, project number 390727645; the German Federal Ministry of Education and Research (BMBF) through the T\"{u}bingen AI Center (FKZ: 01IS18039A); and funds from the Ministry of Science, Research and Arts of the State of Baden-W\"{u}rttemberg. AK is grateful to the International Max Planck Research School for Intelligent Systems (IMPRS-IS) for support. AK is also grateful to Felix Dangel and Jonathan Schmidt for feedback.


  \bibliography{main}
  \bibliographystyle{abbrvnat}

  \clearpage

  \onecolumn \makesupplementtitle
  \thispagestyle{empty}

  \begin{appendices}
    \crefalias{section}{appsec}
    \crefalias{subsection}{appsec}


\section{OOD Test Sets}
\label{sec:app:ood_dataset}

For image-based OOD detection tasks, we use the following test sets on top of MNIST, F-MNIST, SVHN, CIFAR-10, and CIFAR-100:
\begin{itemize}
  \item \textsc{E-MNIST:} Contains handwritten letters (``a''-``z'')---same format as MNIST \citep{cohen2017emnist}.
  \item \textsc{K-MNIST:} Contains handwritten Hiragana scripts---same format as MNIST \citep{clanuwat2018kmnist}.
  \item \textsc{LSUN-CR:} Contains real-world images of classrooms \citep{yu2015lsun}.
  \item \textsc{CIFAR-Gr:} Obtained by converting CIFAR-10 test images to grayscale.
  \item \textsc{F-MNIST-3D:} Obtained by converting single-channel F-MNIST images into three-channel images---all these three channels have identical values.
  \item \textsc{Uniform:} Obtained by drawing independent uniformly-distributed random pixel.
  \item \textsc{Smooth:} Obtained by permuting, smoothing, and contrast-rescaling the original (i.e. the respective in-distribution) test images \citep{hein2019relu}.
\end{itemize}
Meanwhile, for text classification, we use the following OOD test set, following \citep{hendrycks2018deep}:
\begin{itemize}
  \item \textsc{Multi30k:} Multilingual English-German image description dataset \citep{elliot2016multi30k}.
  \item \textsc{WMT16:} Machine-translation dataset, avaliable at \url{http://www.statmt.org/wmt14/translation-task.html}.
  \item \textsc{SNLI:} Collection of human-written English sentence pairs manually labeled for balanced classification with the labels entailment, contradiction, and neutral \citep{bowman2015large}.
\end{itemize}
Finally, for dataset-shift robustness tasks, we use the standard dataset:
\begin{itemize}
  \item \textsc{CIFAR-10-C}: Contains 19 different perturbations---e.g. snow, motion blur, brightness rescaling---with 5 level of severity for a total of 95 distinct shifts \citep{hendrycks2019benchmarking,ovadia2019can}.
\end{itemize}

\section{Training Details}
\label{sec:app:training_details}

\paragraph*{Non-Bayesian}
For MNIST and F-MNIST, we use a five-layer LeNet architecture. Meanwhile, for SVHN, CIFAR-10, and CIFAR-100, we use WideResNet-16-4 \citep{zagoruyko2016wide}.
For all methods, the training procedures are as follows.
For LeNet, we use Adam with initial learning rate of \num{1e-3} and annealed it using the cosine decay method \citep{loshchilov2016sgdr} along with weight decay of \num{5e-4} for 100 epochs.
We use a batch size of 128 for both in- and out-distribution batches, amounting to an effective batch size of 256 in the case of OOD training.
The standard data augmentation pipeline (random crop and horizontal flip) is applied to both in-distribution and OOD data.
For WideResNet-16-4, we use SGD instead with an initial learning rate of \num{1e-1} and Nesterov momentum of 0.9 along with the dropout regularization with rate 0.3---all other hyperparameters are identical to LeNet.
Finally, we use 5 ensemble members for DE.

\paragraph*{Bayesian}
For both LA, VB, and their variants (i.e.\ LA+X and VB+X), we use the identical setup as in the non-Bayesian training above.
Additionally, for LA and LA+X, we use the diagonal Fisher matrix as the approximate Hessian.
Moreover, we tune prior variance by minimizing the validation Brier score.
All predictions are done using 20 MC samples.
For VB and VB+X, we use a diagonal Gaussian variational posterior for both the last-layer weight matrix and bias vector.
Moreover, the prior is a zero-mean isotropic Gaussian with prior precision \num{5e-4} (to emulate the choice of the weight decay in the non-Bayesian training).
The trade-off hyperparameter \(\tau\) of the ELBO is set to the standard value of 0.1 \citep{osawa2019practical,zhang2018noisy}.
We do not use weight decay on the last layer since the regularization of its parameters is done by the KL-term of the ELBO.
Lastly, we use 5 and 200 MC samples for computing the ELBO and for making predictions, respectively.

\paragraph*{Text Classification}
The network used is a two-layer Gated Recurrent Unit \citep[GRU,][]{cho2014learning} with 128 hidden units on each layer.
The word-embedding dimension is 50 and the maximum vocabulary size is 10000.
We put an affine layer on top of the last GRU output to translate the hidden units to output units.
Both the LA and VB are applied only on this layer.
We use a batch size of 64 and Adam optimizer with a learning rate of 0.01 without weight decay, except for LA in which case we use weight decay of \num{5e-4}.
The optimization is done for 5 epochs, following \citep{hendrycks2018deep}.

\section{Additional Results}
\label{sec:app:add_results}

The detailed, non-averaged results for the FPR95 metric are in \cref{tab:fpr95}. Furthermore, additional results with the area-under-ROC-curve (AUROC), area-under-precision-recall-curve (AUPRC), and mean confidence (MMC) metrics are in \cref{tab:fpr95,tab:auroc,tab:auprc,tab:mmc}, respectively. For the full results for models trained with the \textsc{Smooth} noise dataset as $\Dout$ are in \cref{tab:acc_ece_smooth,tab:fpr95_noise_more}. Furthermore, the full results of the NLP experiment is in \cref{tab:acc_ece_nlp,tab:fpr95_nlp_detailed}. Finally, detailed, non-averaged results for sophisticated models (Flipout and CSGHMC) are in \cref{tab:acc_ece_aux,tab:ood_aux_more}.















\begin{table*}[ht]
  \caption{OOD data detection in terms of FPR95. Lower is better. Values are averages over five prediction runs.}
  \label{tab:fpr95}

  \centering

  \resizebox{\textwidth}{!}{
    \begin{tabular}{lrrrrrHrrrrrHrrr}
      \toprule

       & & & & \multicolumn{6}{c}{\bf VB} & \multicolumn{6}{c}{\bf LA} \\
       \cmidrule(r){5-10} \cmidrule(l){11-16}
      \textbf{Datasets} & {\bf MAP} & {\bf OE} & {\bf DE} & {\bf Plain} & {\bf NC} & {\bf NC2} & {\bf SL} & {\bf ML} & {\bf OE} & {\bf Plain} & {\bf NC} & {\bf NC2} & {\bf SL} & {\bf ML} & {\bf OE} \\
      \midrule

      \textbf{MNIST} & & & & & & \\
      F-MNIST & 11.8$\pm$0.0 & 0.0$\pm$0.0 & 5.3$\pm$0.0 & 12.5$\pm$0.1 & 0.1$\pm$0.0 & 0.0$\pm$0.0 & 0.0$\pm$0.0 & 0.4$\pm$0.0 & 1.1$\pm$0.0 & 12.0$\pm$0.0 & 0.2$\pm$0.0 & 0.0$\pm$0.0 & 0.0$\pm$0.0 & 0.1$\pm$0.0 & 0.0$\pm$0.0 \\
      E-MNIST & 35.6$\pm$0.0 & 26.4$\pm$0.0 & 30.4$\pm$0.0 & 34.5$\pm$0.1 & 34.7$\pm$0.1 & 32.8$\pm$0.4 & 14.3$\pm$0.1 & 34.2$\pm$0.1 & 31.4$\pm$0.1 & 35.8$\pm$0.1 & 30.6$\pm$0.0 & 19.5$\pm$0.0 & 12.6$\pm$0.1 & 26.8$\pm$0.1 & 26.7$\pm$0.1 \\
      K-MNIST & 14.4$\pm$0.0 & 5.9$\pm$0.0 & 7.7$\pm$0.0 & 14.0$\pm$0.1 & 10.5$\pm$0.1 & 8.1$\pm$0.1 & 2.1$\pm$0.0 & 9.7$\pm$0.1 & 8.5$\pm$0.0 & 14.5$\pm$0.1 & 8.9$\pm$0.0 & 6.5$\pm$0.0 & 0.7$\pm$0.0 & 5.8$\pm$0.0 & 5.9$\pm$0.0 \\
      CIFAR-Gr & 0.2$\pm$0.0 & 0.0$\pm$0.0 & 0.0$\pm$0.0 & 0.2$\pm$0.0 & 0.0$\pm$0.0 & 0.0$\pm$0.0 & 0.0$\pm$0.0 & 0.0$\pm$0.0 & 0.0$\pm$0.0 & 0.2$\pm$0.0 & 0.0$\pm$0.0 & 0.0$\pm$0.0 & 0.0$\pm$0.0 & 0.0$\pm$0.0 & 0.0$\pm$0.0 \\
      Uniform & 44.3$\pm$0.0 & 0.0$\pm$0.0 & 19.8$\pm$0.0 & 93.1$\pm$0.2 & 0.0$\pm$0.0 & 0.0$\pm$0.0 & 0.0$\pm$0.0 & 0.0$\pm$0.0 & 0.0$\pm$0.0 & 54.2$\pm$0.4 & 0.0$\pm$0.0 & 0.0$\pm$0.0 & 0.0$\pm$0.0 & 0.0$\pm$0.0 & 0.0$\pm$0.0 \\
      Smooth & 0.0$\pm$0.0 & 0.0$\pm$0.0 & 0.0$\pm$0.0 & 0.0$\pm$0.0 & 0.0$\pm$0.0 & 0.0$\pm$0.0 & 0.0$\pm$0.0 & 0.0$\pm$0.0 & 0.0$\pm$0.0 & 0.0$\pm$0.0 & 0.0$\pm$0.0 & 0.0$\pm$0.0 & 0.0$\pm$0.0 & 0.0$\pm$0.0 & 0.0$\pm$0.0 \\

      \midrule

      \textbf{F-MNIST} & & & & & & \\
      MNIST & 73.5$\pm$0.0 & 38.5$\pm$0.0 & 65.8$\pm$0.0 & 66.8$\pm$0.1 & 43.5$\pm$0.1 & 10.4$\pm$0.1 & 9.5$\pm$0.0 & 50.1$\pm$0.1 & 57.2$\pm$0.1 & 72.2$\pm$0.2 & 25.6$\pm$0.1 & 3.5$\pm$0.0 & 11.5$\pm$0.1 & 38.9$\pm$0.3 & 39.9$\pm$0.1 \\
      E-MNIST & 73.6$\pm$0.0 & 21.0$\pm$0.0 & 58.6$\pm$0.0 & 68.1$\pm$0.1 & 18.7$\pm$0.0 & 3.0$\pm$0.0 & 5.0$\pm$0.0 & 34.0$\pm$0.0 & 40.6$\pm$0.0 & 72.2$\pm$0.2 & 6.0$\pm$0.0 & 0.5$\pm$0.0 & 4.6$\pm$0.1 & 14.7$\pm$0.1 & 23.1$\pm$0.1 \\
      K-MNIST & 73.7$\pm$0.0 & 37.4$\pm$0.0 & 47.2$\pm$0.0 & 62.6$\pm$0.1 & 28.0$\pm$0.1 & 6.1$\pm$0.1 & 10.6$\pm$0.0 & 33.4$\pm$0.0 & 36.7$\pm$0.0 & 71.5$\pm$0.2 & 18.2$\pm$0.1 & 1.8$\pm$0.0 & 8.7$\pm$0.1 & 32.5$\pm$0.2 & 38.7$\pm$0.3 \\
      CIFAR-Gr & 87.2$\pm$0.0 & 0.0$\pm$0.0 & 86.6$\pm$0.0 & 75.3$\pm$0.0 & 0.0$\pm$0.0 & 0.0$\pm$0.0 & 0.0$\pm$0.0 & 0.0$\pm$0.0 & 0.0$\pm$0.0 & 87.7$\pm$0.1 & 0.0$\pm$0.0 & 0.0$\pm$0.0 & 0.0$\pm$0.0 & 0.0$\pm$0.0 & 0.0$\pm$0.0 \\
      Uniform & 81.3$\pm$0.0 & 0.0$\pm$0.0 & 86.3$\pm$0.0 & 87.3$\pm$0.1 & 0.0$\pm$0.0 & 0.0$\pm$0.0 & 0.0$\pm$0.0 & 0.0$\pm$0.0 & 0.0$\pm$0.0 & 81.0$\pm$0.2 & 0.0$\pm$0.0 & 0.0$\pm$0.0 & 0.0$\pm$0.0 & 0.0$\pm$0.0 & 0.1$\pm$0.0 \\
      Smooth & 26.8$\pm$0.0 & 0.0$\pm$0.0 & 24.2$\pm$0.0 & 19.6$\pm$0.0 & 0.0$\pm$0.0 & 0.0$\pm$0.0 & 0.0$\pm$0.0 & 0.2$\pm$0.0 & 0.1$\pm$0.0 & 27.3$\pm$0.1 & 0.0$\pm$0.0 & 0.0$\pm$0.0 & 0.0$\pm$0.0 & 0.0$\pm$0.0 & 0.0$\pm$0.0 \\

      \midrule

      \textbf{SVHN} & & & & & & \\
      CIFAR-10 & 18.9$\pm$0.0 & 0.1$\pm$0.0 & 9.5$\pm$0.0 & 15.0$\pm$0.0 & 0.3$\pm$0.0 & 0.1$\pm$0.0 & 0.1$\pm$0.0 & 0.0$\pm$0.0 & 0.1$\pm$0.0 & 15.4$\pm$0.1 & 0.4$\pm$0.0 & 0.0$\pm$0.0 & 0.0$\pm$0.0 & 0.0$\pm$0.0 & 0.1$\pm$0.0 \\
      LSUN-CR & 19.7$\pm$0.0 & 0.0$\pm$0.0 & 8.3$\pm$0.0 & 17.2$\pm$0.2 & 0.0$\pm$0.0 & 0.0$\pm$0.0 & 0.0$\pm$0.0 & 0.0$\pm$0.0 & 0.0$\pm$0.0 & 15.5$\pm$0.1 & 0.0$\pm$0.0 & 0.0$\pm$0.0 & 0.0$\pm$0.0 & 0.0$\pm$0.0 & 0.0$\pm$0.0 \\
      CIFAR-100 & 21.8$\pm$0.0 & 0.2$\pm$0.0 & 11.6$\pm$0.0 & 18.1$\pm$0.0 & 0.5$\pm$0.0 & 0.2$\pm$0.0 & 0.5$\pm$0.0 & 0.1$\pm$0.0 & 0.2$\pm$0.0 & 17.6$\pm$0.1 & 0.6$\pm$0.0 & 0.1$\pm$0.0 & 0.2$\pm$0.0 & 0.2$\pm$0.0 & 0.1$\pm$0.0 \\
      FMNIST-3D & 26.7$\pm$0.0 & 0.0$\pm$0.0 & 17.5$\pm$0.0 & 24.5$\pm$0.1 & 0.0$\pm$0.0 & 0.0$\pm$0.0 & 0.6$\pm$0.0 & 0.0$\pm$0.0 & 0.0$\pm$0.0 & 27.2$\pm$0.1 & 0.1$\pm$0.0 & 0.0$\pm$0.0 & 0.0$\pm$0.0 & 0.0$\pm$0.0 & 0.0$\pm$0.0 \\
      Uniform & 30.0$\pm$0.0 & 0.0$\pm$0.0 & 6.4$\pm$0.0 & 48.2$\pm$0.0 & 0.0$\pm$0.0 & 0.0$\pm$0.0 & 0.0$\pm$0.0 & 0.0$\pm$0.0 & 0.0$\pm$0.0 & 17.0$\pm$0.1 & 0.0$\pm$0.0 & 0.0$\pm$0.0 & 0.0$\pm$0.0 & 0.0$\pm$0.0 & 0.0$\pm$0.0 \\
      Smooth & 17.3$\pm$0.0 & 12.0$\pm$0.0 & 6.9$\pm$0.0 & 9.1$\pm$0.0 & 7.7$\pm$0.0 & 9.7$\pm$0.1 & 9.5$\pm$0.0 & 8.3$\pm$0.0 & 8.4$\pm$0.0 & 10.1$\pm$0.1 & 8.1$\pm$0.1 & 5.3$\pm$0.1 & 5.9$\pm$0.0 & 6.6$\pm$0.1 & 6.4$\pm$0.0 \\

      \midrule

      \textbf{CIFAR-10} & & & & & & \\
      SVHN & 34.5$\pm$0.0 & 10.0$\pm$0.0 & 33.9$\pm$0.0 & 33.5$\pm$0.0 & 30.6$\pm$0.1 & 5.3$\pm$0.0 & 59.4$\pm$0.0 & 18.3$\pm$0.1 & 33.9$\pm$0.1 & 35.5$\pm$0.1 & 12.7$\pm$0.2 & 2.6$\pm$0.0 & 47.2$\pm$0.3 & 8.7$\pm$0.1 & 10.8$\pm$0.0 \\
      LSUN-CR & 53.3$\pm$0.0 & 28.0$\pm$0.0 & 44.0$\pm$0.0 & 49.4$\pm$0.4 & 25.9$\pm$0.2 & 8.3$\pm$0.1 & 43.7$\pm$0.1 & 36.8$\pm$0.1 & 34.8$\pm$0.2 & 53.8$\pm$0.6 & 17.5$\pm$0.3 & 1.3$\pm$0.1 & 41.2$\pm$0.9 & 30.1$\pm$0.5 & 28.4$\pm$0.4 \\
      CIFAR-100 & 61.2$\pm$0.0 & 57.8$\pm$0.0 & 52.5$\pm$0.0 & 58.4$\pm$0.1 & 58.5$\pm$0.1 & 55.3$\pm$0.1 & 63.3$\pm$0.0 & 56.8$\pm$0.1 & 57.1$\pm$0.1 & 61.4$\pm$0.1 & 59.6$\pm$0.2 & 62.6$\pm$0.2 & 62.2$\pm$0.2 & 60.4$\pm$0.2 & 57.9$\pm$0.1 \\
      FMNIST-3D & 42.4$\pm$0.0 & 26.8$\pm$0.0 & 30.7$\pm$0.0 & 37.4$\pm$0.0 & 19.0$\pm$0.1 & 5.4$\pm$0.0 & 43.9$\pm$0.0 & 32.2$\pm$0.1 & 29.6$\pm$0.1 & 43.2$\pm$0.2 & 15.4$\pm$0.1 & 2.8$\pm$0.0 & 36.8$\pm$0.2 & 24.2$\pm$0.1 & 27.8$\pm$0.1 \\
      Uniform & 87.7$\pm$0.0 & 0.0$\pm$0.0 & 0.0$\pm$0.0 & 13.8$\pm$0.1 & 0.0$\pm$0.0 & 0.0$\pm$0.0 & 0.0$\pm$0.0 & 0.0$\pm$0.0 & 0.0$\pm$0.0 & 92.8$\pm$0.1 & 0.0$\pm$0.0 & 0.0$\pm$0.0 & 0.0$\pm$0.0 & 0.0$\pm$0.0 & 0.0$\pm$0.0 \\
      Smooth & 35.1$\pm$0.0 & 14.2$\pm$0.0 & 32.9$\pm$0.0 & 26.4$\pm$0.0 & 34.0$\pm$0.2 & 8.1$\pm$0.0 & 31.9$\pm$0.0 & 30.3$\pm$0.0 & 23.1$\pm$0.1 & 34.9$\pm$0.1 & 15.5$\pm$0.2 & 2.8$\pm$0.0 & 43.6$\pm$0.2 & 7.5$\pm$0.0 & 14.9$\pm$0.1 \\

      \midrule

      \textbf{CIFAR-100} & & & & & & \\
      LSUN-CR & 82.0$\pm$0.0 & 64.3$\pm$0.0 & 75.3$\pm$0.0 & 73.8$\pm$0.2 & 62.3$\pm$0.5 & 23.9$\pm$0.6 & 76.3$\pm$0.1 & 65.3$\pm$0.1 & 67.6$\pm$0.3 & 82.8$\pm$0.5 & 55.9$\pm$0.5 & 13.5$\pm$0.2 & 75.6$\pm$0.7 & 65.3$\pm$0.5 & 64.1$\pm$1.0 \\
      CIFAR-10 & 79.8$\pm$0.0 & 81.9$\pm$0.0 & 76.4$\pm$0.0 & 78.2$\pm$0.1 & 81.4$\pm$0.1 & 90.9$\pm$0.2 & 82.8$\pm$0.0 & 79.5$\pm$0.1 & 79.0$\pm$0.0 & 79.5$\pm$0.1 & 80.9$\pm$0.2 & 91.4$\pm$0.0 & 81.7$\pm$0.1 & 80.8$\pm$0.1 & 80.0$\pm$0.1 \\
      FMNIST-3D & 65.8$\pm$0.0 & 58.5$\pm$0.0 & 61.8$\pm$0.0 & 57.1$\pm$0.1 & 41.0$\pm$0.2 & 14.0$\pm$0.2 & 72.0$\pm$0.1 & 51.7$\pm$0.1 & 56.0$\pm$0.1 & 66.1$\pm$0.1 & 58.6$\pm$0.3 & 21.1$\pm$0.1 & 69.0$\pm$0.2 & 59.2$\pm$0.1 & 59.3$\pm$0.3 \\
      Uniform & 97.6$\pm$0.0 & 0.0$\pm$0.0 & 94.3$\pm$0.0 & 100.0$\pm$0.0 & 0.0$\pm$0.0 & 0.0$\pm$0.0 & 0.0$\pm$0.0 & 0.0$\pm$0.0 & 0.0$\pm$0.0 & 98.8$\pm$0.1 & 0.0$\pm$0.0 & 0.0$\pm$0.0 & 0.0$\pm$0.0 & 0.1$\pm$0.0 & 0.0$\pm$0.0 \\
      Smooth & 79.5$\pm$0.0 & 65.2$\pm$0.0 & 58.7$\pm$0.0 & 79.1$\pm$0.0 & 64.8$\pm$0.1 & 29.4$\pm$0.1 & 80.2$\pm$0.1 & 54.4$\pm$0.1 & 64.0$\pm$0.0 & 79.2$\pm$0.1 & 41.6$\pm$0.2 & 2.6$\pm$0.0 & 78.0$\pm$0.1 & 57.1$\pm$0.3 & 66.2$\pm$0.1 \\

      \bottomrule
    \end{tabular}
  }
\end{table*}

\begin{table*}[ht]
  \caption{OOD data detection in terms of AUROC. Higher is better. Values are averages over five prediction runs.}
  \label{tab:auroc}

  \centering

  \resizebox{\textwidth}{!}{
    \begin{tabular}{lrrrrrHrrrrrHrrr}
      \toprule

       & & & & \multicolumn{6}{c}{\bf VB} & \multicolumn{6}{c}{\bf LA} \\
       \cmidrule(r){5-10} \cmidrule(l){11-16}
      \textbf{Datasets} & {\bf MAP} & {\bf OE} & {\bf DE} & {\bf Plain} & {\bf NC} & {\bf NC2} & {\bf SL} & {\bf ML} & {\bf OE} & {\bf Plain} & {\bf NC} & {\bf NC2} & {\bf SL} & {\bf ML} & {\bf OE} \\
      \midrule

      \textbf{MNIST} & & & & & & \\
      F-MNIST & 97.3$\pm$0.0 & 99.9$\pm$0.0 & 98.7$\pm$0.0 & 97.4$\pm$0.0 & 99.9$\pm$0.0 & 100.0$\pm$0.0 & 99.9$\pm$0.0 & 99.8$\pm$0.0 & 99.6$\pm$0.0 & 97.4$\pm$0.0 & 99.9$\pm$0.0 & 100.0$\pm$0.0 & 99.9$\pm$0.0 & 99.9$\pm$0.0 & 99.9$\pm$0.0 \\
      E-MNIST & 89.1$\pm$0.0 & 93.7$\pm$0.0 & 90.4$\pm$0.0 & 89.9$\pm$0.1 & 90.4$\pm$0.0 & 90.7$\pm$0.3 & 95.7$\pm$0.0 & 91.1$\pm$0.0 & 92.1$\pm$0.1 & 89.1$\pm$0.0 & 91.2$\pm$0.0 & 95.6$\pm$0.0 & 94.9$\pm$0.0 & 93.3$\pm$0.0 & 93.6$\pm$0.0 \\
      K-MNIST & 96.9$\pm$0.0 & 98.5$\pm$0.0 & 98.1$\pm$0.0 & 96.9$\pm$0.0 & 97.8$\pm$0.0 & 98.3$\pm$0.0 & 98.8$\pm$0.0 & 98.0$\pm$0.0 & 98.2$\pm$0.0 & 96.9$\pm$0.0 & 97.9$\pm$0.0 & 98.6$\pm$0.0 & 99.2$\pm$0.0 & 98.5$\pm$0.0 & 98.5$\pm$0.0 \\
      CIFAR-Gr & 99.6$\pm$0.0 & 100.0$\pm$0.0 & 99.8$\pm$0.0 & 99.6$\pm$0.0 & 100.0$\pm$0.0 & 100.0$\pm$0.0 & 100.0$\pm$0.0 & 100.0$\pm$0.0 & 100.0$\pm$0.0 & 99.6$\pm$0.0 & 100.0$\pm$0.0 & 100.0$\pm$0.0 & 100.0$\pm$0.0 & 100.0$\pm$0.0 & 100.0$\pm$0.0 \\
      Uniform & 95.0$\pm$0.0 & 100.0$\pm$0.0 & 95.8$\pm$0.0 & 90.5$\pm$0.0 & 100.0$\pm$0.0 & 100.0$\pm$0.0 & 100.0$\pm$0.0 & 100.0$\pm$0.0 & 100.0$\pm$0.0 & 94.6$\pm$0.0 & 100.0$\pm$0.0 & 100.0$\pm$0.0 & 100.0$\pm$0.0 & 100.0$\pm$0.0 & 100.0$\pm$0.0 \\
      Smooth & 100.0$\pm$0.0 & 100.0$\pm$0.0 & 100.0$\pm$0.0 & 100.0$\pm$0.0 & 100.0$\pm$0.0 & 100.0$\pm$0.0 & 100.0$\pm$0.0 & 100.0$\pm$0.0 & 100.0$\pm$0.0 & 100.0$\pm$0.0 & 100.0$\pm$0.0 & 100.0$\pm$0.0 & 100.0$\pm$0.0 & 100.0$\pm$0.0 & 100.0$\pm$0.0 \\

      \midrule

      \textbf{F-MNIST} & & & & & & \\
      MNIST & 79.7$\pm$0.0 & 92.9$\pm$0.0 & 83.0$\pm$0.0 & 85.3$\pm$0.0 & 87.9$\pm$0.0 & 96.9$\pm$0.0 & 98.6$\pm$0.0 & 86.7$\pm$0.0 & 86.2$\pm$0.0 & 80.3$\pm$0.0 & 94.2$\pm$0.0 & 99.3$\pm$0.0 & 98.2$\pm$0.0 & 92.9$\pm$0.0 & 92.5$\pm$0.0 \\
      E-MNIST & 81.8$\pm$0.0 & 96.5$\pm$0.0 & 87.5$\pm$0.0 & 85.1$\pm$0.0 & 95.6$\pm$0.0 & 99.2$\pm$0.0 & 99.2$\pm$0.0 & 92.0$\pm$0.0 & 91.3$\pm$0.0 & 82.3$\pm$0.0 & 98.9$\pm$0.0 & 99.9$\pm$0.0 & 99.3$\pm$0.0 & 97.6$\pm$0.0 & 96.1$\pm$0.0 \\
      K-MNIST & 83.1$\pm$0.0 & 94.4$\pm$0.0 & 91.7$\pm$0.0 & 86.9$\pm$0.0 & 94.3$\pm$0.0 & 98.6$\pm$0.0 & 98.4$\pm$0.0 & 93.5$\pm$0.0 & 93.3$\pm$0.0 & 83.9$\pm$0.0 & 96.9$\pm$0.0 & 99.5$\pm$0.0 & 98.7$\pm$0.0 & 94.9$\pm$0.0 & 94.1$\pm$0.0 \\
      CIFAR-Gr & 82.2$\pm$0.0 & 100.0$\pm$0.0 & 83.6$\pm$0.0 & 87.5$\pm$0.0 & 100.0$\pm$0.0 & 100.0$\pm$0.0 & 100.0$\pm$0.0 & 100.0$\pm$0.0 & 100.0$\pm$0.0 & 81.4$\pm$0.0 & 100.0$\pm$0.0 & 100.0$\pm$0.0 & 100.0$\pm$0.0 & 100.0$\pm$0.0 & 100.0$\pm$0.0 \\
      Uniform & 85.5$\pm$0.0 & 100.0$\pm$0.0 & 85.7$\pm$0.0 & 85.8$\pm$0.0 & 100.0$\pm$0.0 & 100.0$\pm$0.0 & 100.0$\pm$0.0 & 100.0$\pm$0.0 & 100.0$\pm$0.0 & 85.3$\pm$0.0 & 100.0$\pm$0.0 & 100.0$\pm$0.0 & 100.0$\pm$0.0 & 100.0$\pm$0.0 & 100.0$\pm$0.0 \\
      Smooth & 95.7$\pm$0.0 & 100.0$\pm$0.0 & 96.4$\pm$0.0 & 97.2$\pm$0.0 & 100.0$\pm$0.0 & 100.0$\pm$0.0 & 100.0$\pm$0.0 & 100.0$\pm$0.0 & 100.0$\pm$0.0 & 95.5$\pm$0.0 & 100.0$\pm$0.0 & 100.0$\pm$0.0 & 100.0$\pm$0.0 & 100.0$\pm$0.0 & 100.0$\pm$0.0 \\

      \midrule

      \textbf{SVHN} & & & & & & \\
      CIFAR-10 & 96.2$\pm$0.0 & 100.0$\pm$0.0 & 97.9$\pm$0.0 & 95.6$\pm$0.0 & 99.9$\pm$0.0 & 100.0$\pm$0.0 & 99.9$\pm$0.0 & 100.0$\pm$0.0 & 100.0$\pm$0.0 & 97.1$\pm$0.0 & 99.9$\pm$0.0 & 100.0$\pm$0.0 & 100.0$\pm$0.0 & 100.0$\pm$0.0 & 100.0$\pm$0.0 \\
      LSUN-CR & 95.7$\pm$0.0 & 100.0$\pm$0.0 & 97.7$\pm$0.0 & 95.9$\pm$0.0 & 100.0$\pm$0.0 & 100.0$\pm$0.0 & 100.0$\pm$0.0 & 100.0$\pm$0.0 & 100.0$\pm$0.0 & 97.0$\pm$0.0 & 100.0$\pm$0.0 & 100.0$\pm$0.0 & 100.0$\pm$0.0 & 100.0$\pm$0.0 & 100.0$\pm$0.0 \\
      CIFAR-100 & 95.5$\pm$0.0 & 99.9$\pm$0.0 & 97.4$\pm$0.0 & 94.7$\pm$0.0 & 99.9$\pm$0.0 & 99.9$\pm$0.0 & 99.8$\pm$0.0 & 100.0$\pm$0.0 & 99.9$\pm$0.0 & 96.5$\pm$0.0 & 99.9$\pm$0.0 & 100.0$\pm$0.0 & 99.9$\pm$0.0 & 100.0$\pm$0.0 & 100.0$\pm$0.0 \\
      FMNIST-3D & 95.5$\pm$0.0 & 100.0$\pm$0.0 & 97.1$\pm$0.0 & 91.4$\pm$0.0 & 100.0$\pm$0.0 & 100.0$\pm$0.0 & 99.8$\pm$0.0 & 100.0$\pm$0.0 & 100.0$\pm$0.0 & 95.6$\pm$0.0 & 100.0$\pm$0.0 & 100.0$\pm$0.0 & 99.9$\pm$0.0 & 100.0$\pm$0.0 & 100.0$\pm$0.0 \\
      Uniform & 94.3$\pm$0.0 & 100.0$\pm$0.0 & 98.2$\pm$0.0 & 80.2$\pm$0.0 & 100.0$\pm$0.0 & 100.0$\pm$0.0 & 100.0$\pm$0.0 & 100.0$\pm$0.0 & 100.0$\pm$0.0 & 96.8$\pm$0.0 & 100.0$\pm$0.0 & 100.0$\pm$0.0 & 100.0$\pm$0.0 & 100.0$\pm$0.0 & 100.0$\pm$0.0 \\
      Smooth & 96.5$\pm$0.0 & 97.6$\pm$0.0 & 98.4$\pm$0.0 & 97.5$\pm$0.0 & 97.7$\pm$0.0 & 97.0$\pm$0.0 & 95.9$\pm$0.0 & 97.5$\pm$0.0 & 97.7$\pm$0.0 & 97.7$\pm$0.0 & 98.4$\pm$0.0 & 98.8$\pm$0.0 & 98.1$\pm$0.0 & 98.5$\pm$0.0 & 98.7$\pm$0.0 \\

      \midrule

      \textbf{CIFAR-10} & & & & & & \\
      SVHN & 95.6$\pm$0.0 & 98.2$\pm$0.0 & 95.6$\pm$0.0 & 95.7$\pm$0.0 & 95.8$\pm$0.0 & 98.2$\pm$0.0 & 89.0$\pm$0.0 & 97.2$\pm$0.0 & 95.6$\pm$0.0 & 95.5$\pm$0.0 & 97.8$\pm$0.0 & 98.6$\pm$0.0 & 92.7$\pm$0.0 & 98.7$\pm$0.0 & 98.1$\pm$0.0 \\
      LSUN-CR & 91.8$\pm$0.0 & 95.9$\pm$0.0 & 93.7$\pm$0.0 & 91.5$\pm$0.0 & 96.1$\pm$0.0 & 98.0$\pm$0.0 & 92.5$\pm$0.0 & 94.1$\pm$0.0 & 94.4$\pm$0.0 & 92.0$\pm$0.0 & 97.5$\pm$0.0 & 99.1$\pm$0.0 & 93.8$\pm$0.1 & 95.7$\pm$0.0 & 96.0$\pm$0.1 \\
      CIFAR-100 & 89.8$\pm$0.0 & 90.1$\pm$0.0 & 91.3$\pm$0.0 & 88.6$\pm$0.0 & 88.2$\pm$0.0 & 87.6$\pm$0.0 & 85.9$\pm$0.0 & 88.3$\pm$0.0 & 88.9$\pm$0.0 & 89.9$\pm$0.0 & 89.8$\pm$0.0 & 89.0$\pm$0.0 & 86.2$\pm$0.0 & 89.7$\pm$0.0 & 90.0$\pm$0.0 \\
      FMNIST-3D & 94.4$\pm$0.0 & 96.2$\pm$0.0 & 95.8$\pm$0.0 & 94.5$\pm$0.0 & 97.2$\pm$0.0 & 98.6$\pm$0.0 & 92.9$\pm$0.0 & 95.1$\pm$0.0 & 95.9$\pm$0.0 & 94.3$\pm$0.0 & 97.7$\pm$0.0 & 98.9$\pm$0.0 & 94.0$\pm$0.0 & 96.6$\pm$0.0 & 96.1$\pm$0.0 \\
      Uniform & 93.0$\pm$0.0 & 100.0$\pm$0.0 & 99.5$\pm$0.0 & 97.6$\pm$0.0 & 100.0$\pm$0.0 & 100.0$\pm$0.0 & 100.0$\pm$0.0 & 100.0$\pm$0.0 & 100.0$\pm$0.0 & 92.2$\pm$0.0 & 100.0$\pm$0.0 & 100.0$\pm$0.0 & 100.0$\pm$0.0 & 100.0$\pm$0.0 & 100.0$\pm$0.0 \\
      Smooth & 94.3$\pm$0.0 & 97.6$\pm$0.0 & 95.6$\pm$0.0 & 96.0$\pm$0.0 & 95.3$\pm$0.0 & 97.5$\pm$0.0 & 94.8$\pm$0.0 & 95.4$\pm$0.0 & 96.2$\pm$0.0 & 94.6$\pm$0.0 & 97.6$\pm$0.0 & 98.6$\pm$0.0 & 93.9$\pm$0.0 & 98.8$\pm$0.0 & 97.6$\pm$0.0 \\

      \midrule

      \textbf{CIFAR-100} & & & & & & \\
      LSUN-CR & 78.4$\pm$0.0 & 85.3$\pm$0.0 & 83.8$\pm$0.0 & 81.3$\pm$0.1 & 87.9$\pm$0.0 & 93.0$\pm$0.1 & 80.1$\pm$0.0 & 85.6$\pm$0.0 & 85.3$\pm$0.1 & 78.8$\pm$0.1 & 89.1$\pm$0.0 & 96.9$\pm$0.0 & 82.0$\pm$0.3 & 86.3$\pm$0.1 & 85.9$\pm$0.2 \\
      CIFAR-10 & 77.4$\pm$0.0 & 77.1$\pm$0.0 & 79.8$\pm$0.0 & 77.7$\pm$0.0 & 76.8$\pm$0.0 & 74.1$\pm$0.0 & 76.3$\pm$0.0 & 77.8$\pm$0.0 & 77.3$\pm$0.0 & 77.8$\pm$0.0 & 77.5$\pm$0.0 & 70.1$\pm$0.0 & 76.6$\pm$0.0 & 77.6$\pm$0.0 & 77.7$\pm$0.0 \\
      FMNIST-3D & 85.3$\pm$0.0 & 86.2$\pm$0.0 & 87.6$\pm$0.0 & 87.7$\pm$0.0 & 91.5$\pm$0.0 & 96.6$\pm$0.0 & 84.0$\pm$0.0 & 89.6$\pm$0.0 & 87.9$\pm$0.0 & 85.2$\pm$0.0 & 87.0$\pm$0.0 & 96.3$\pm$0.0 & 85.7$\pm$0.0 & 86.3$\pm$0.0 & 86.0$\pm$0.0 \\
      Uniform & 80.1$\pm$0.0 & 100.0$\pm$0.0 & 87.7$\pm$0.0 & 64.8$\pm$0.0 & 100.0$\pm$0.0 & 100.0$\pm$0.0 & 100.0$\pm$0.0 & 100.0$\pm$0.0 & 100.0$\pm$0.0 & 82.6$\pm$0.1 & 100.0$\pm$0.0 & 100.0$\pm$0.0 & 100.0$\pm$0.0 & 99.9$\pm$0.0 & 100.0$\pm$0.0 \\
      Smooth & 77.0$\pm$0.0 & 76.6$\pm$0.0 & 82.9$\pm$0.0 & 69.4$\pm$0.0 & 74.5$\pm$0.0 & 90.5$\pm$0.0 & 80.3$\pm$0.0 & 83.2$\pm$0.0 & 78.3$\pm$0.0 & 79.4$\pm$0.1 & 91.7$\pm$0.0 & 98.7$\pm$0.0 & 78.1$\pm$0.0 & 86.4$\pm$0.1 & 77.7$\pm$0.1 \\

      \bottomrule
    \end{tabular}
  }
\end{table*}

\begin{table*}[ht]
  \caption{OOD data detection in terms of AUPRC. Higher is better. Values are averages over five prediction runs.}
  \label{tab:auprc}

  \centering

  \resizebox{\textwidth}{!}{
    \begin{tabular}{lrrrrrHrrrrrHrrr}
      \toprule

       & & & & \multicolumn{6}{c}{\bf VB} & \multicolumn{6}{c}{\bf LA} \\
       \cmidrule(r){5-10} \cmidrule(l){11-16}
      \textbf{Datasets} & {\bf MAP} & {\bf OE} & {\bf DE} & {\bf Plain} & {\bf NC} & {\bf NC2} & {\bf SL} & {\bf ML} & {\bf OE} & {\bf Plain} & {\bf NC} & {\bf NC2} & {\bf SL} & {\bf ML} & {\bf OE} \\
      \midrule

      \textbf{MNIST} & & & & & & \\
      F-MNIST & 96.9$\pm$0.0 & 99.9$\pm$0.0 & 98.7$\pm$0.0 & 97.5$\pm$0.0 & 99.9$\pm$0.0 & 100.0$\pm$0.0 & 99.8$\pm$0.0 & 99.8$\pm$0.0 & 99.6$\pm$0.0 & 97.0$\pm$0.0 & 99.9$\pm$0.0 & 100.0$\pm$0.0 & 99.9$\pm$0.0 & 99.9$\pm$0.0 & 99.9$\pm$0.0 \\
      E-MNIST & 74.2$\pm$0.0 & 86.7$\pm$0.0 & 77.2$\pm$0.0 & 76.3$\pm$0.2 & 77.8$\pm$0.1 & 77.3$\pm$0.8 & 83.3$\pm$0.0 & 80.6$\pm$0.2 & 82.5$\pm$0.2 & 74.1$\pm$0.1 & 79.6$\pm$0.0 & 89.8$\pm$0.0 & 78.8$\pm$0.1 & 85.5$\pm$0.0 & 86.4$\pm$0.0 \\
      K-MNIST & 96.5$\pm$0.0 & 98.5$\pm$0.0 & 98.0$\pm$0.0 & 96.4$\pm$0.0 & 97.5$\pm$0.0 & 97.8$\pm$0.1 & 97.2$\pm$0.0 & 97.8$\pm$0.0 & 98.1$\pm$0.0 & 96.6$\pm$0.0 & 97.8$\pm$0.0 & 98.2$\pm$0.0 & 98.9$\pm$0.0 & 98.4$\pm$0.0 & 98.4$\pm$0.0 \\
      CIFAR-Gr & 99.7$\pm$0.0 & 100.0$\pm$0.0 & 99.8$\pm$0.0 & 99.6$\pm$0.0 & 100.0$\pm$0.0 & 100.0$\pm$0.0 & 100.0$\pm$0.0 & 100.0$\pm$0.0 & 100.0$\pm$0.0 & 99.6$\pm$0.0 & 100.0$\pm$0.0 & 100.0$\pm$0.0 & 100.0$\pm$0.0 & 100.0$\pm$0.0 & 100.0$\pm$0.0 \\
      Uniform & 96.7$\pm$0.0 & 100.0$\pm$0.0 & 97.3$\pm$0.0 & 93.4$\pm$0.0 & 100.0$\pm$0.0 & 100.0$\pm$0.0 & 100.0$\pm$0.0 & 100.0$\pm$0.0 & 100.0$\pm$0.0 & 96.5$\pm$0.0 & 100.0$\pm$0.0 & 100.0$\pm$0.0 & 100.0$\pm$0.0 & 100.0$\pm$0.0 & 100.0$\pm$0.0 \\
      Smooth & 100.0$\pm$0.0 & 100.0$\pm$0.0 & 100.0$\pm$0.0 & 100.0$\pm$0.0 & 100.0$\pm$0.0 & 100.0$\pm$0.0 & 100.0$\pm$0.0 & 100.0$\pm$0.0 & 100.0$\pm$0.0 & 100.0$\pm$0.0 & 100.0$\pm$0.0 & 100.0$\pm$0.0 & 100.0$\pm$0.0 & 100.0$\pm$0.0 & 100.0$\pm$0.0 \\

      \midrule

      \textbf{F-MNIST} & & & & & & \\
      MNIST & 75.3$\pm$0.0 & 92.2$\pm$0.0 & 79.0$\pm$0.0 & 83.1$\pm$0.1 & 84.0$\pm$0.1 & 94.9$\pm$0.1 & 98.4$\pm$0.0 & 83.9$\pm$0.1 & 84.1$\pm$0.1 & 76.3$\pm$0.0 & 92.5$\pm$0.0 & 99.1$\pm$0.0 & 98.0$\pm$0.0 & 92.3$\pm$0.0 & 91.9$\pm$0.0 \\
      E-MNIST & 66.9$\pm$0.0 & 92.7$\pm$0.0 & 76.4$\pm$0.0 & 74.0$\pm$0.3 & 88.7$\pm$0.1 & 97.3$\pm$0.0 & 98.2$\pm$0.0 & 82.8$\pm$0.0 & 82.5$\pm$0.1 & 67.8$\pm$0.0 & 96.8$\pm$0.0 & 99.6$\pm$0.0 & 98.4$\pm$0.0 & 94.8$\pm$0.0 & 92.0$\pm$0.0 \\
      K-MNIST & 81.7$\pm$0.0 & 94.4$\pm$0.0 & 91.1$\pm$0.0 & 85.2$\pm$0.1 & 93.1$\pm$0.0 & 98.1$\pm$0.0 & 98.1$\pm$0.0 & 92.5$\pm$0.0 & 92.6$\pm$0.0 & 82.8$\pm$0.0 & 96.2$\pm$0.0 & 99.5$\pm$0.0 & 98.6$\pm$0.0 & 94.7$\pm$0.0 & 94.2$\pm$0.0 \\
      CIFAR-Gr & 85.5$\pm$0.0 & 100.0$\pm$0.0 & 87.2$\pm$0.0 & 89.6$\pm$0.0 & 100.0$\pm$0.0 & 100.0$\pm$0.0 & 100.0$\pm$0.0 & 100.0$\pm$0.0 & 100.0$\pm$0.0 & 84.9$\pm$0.0 & 100.0$\pm$0.0 & 100.0$\pm$0.0 & 100.0$\pm$0.0 & 100.0$\pm$0.0 & 100.0$\pm$0.0 \\
      Uniform & 88.1$\pm$0.0 & 100.0$\pm$0.0 & 88.9$\pm$0.0 & 89.2$\pm$0.0 & 100.0$\pm$0.0 & 100.0$\pm$0.0 & 100.0$\pm$0.0 & 100.0$\pm$0.0 & 100.0$\pm$0.0 & 87.9$\pm$0.0 & 100.0$\pm$0.0 & 100.0$\pm$0.0 & 100.0$\pm$0.0 & 100.0$\pm$0.0 & 100.0$\pm$0.0 \\
      Smooth & 95.3$\pm$0.0 & 100.0$\pm$0.0 & 96.1$\pm$0.0 & 96.9$\pm$0.0 & 100.0$\pm$0.0 & 100.0$\pm$0.0 & 100.0$\pm$0.0 & 100.0$\pm$0.0 & 100.0$\pm$0.0 & 95.1$\pm$0.0 & 100.0$\pm$0.0 & 100.0$\pm$0.0 & 100.0$\pm$0.0 & 100.0$\pm$0.0 & 100.0$\pm$0.0 \\

      \midrule

      \textbf{SVHN} & & & & & & \\
      CIFAR-10 & 98.3$\pm$0.0 & 100.0$\pm$0.0 & 99.1$\pm$0.0 & 96.9$\pm$0.0 & 99.9$\pm$0.0 & 100.0$\pm$0.0 & 99.9$\pm$0.0 & 100.0$\pm$0.0 & 100.0$\pm$0.0 & 98.8$\pm$0.0 & 100.0$\pm$0.0 & 100.0$\pm$0.0 & 100.0$\pm$0.0 & 100.0$\pm$0.0 & 100.0$\pm$0.0 \\
      LSUN-CR & 99.9$\pm$0.0 & 100.0$\pm$0.0 & 100.0$\pm$0.0 & 99.9$\pm$0.0 & 100.0$\pm$0.0 & 100.0$\pm$0.0 & 100.0$\pm$0.0 & 100.0$\pm$0.0 & 100.0$\pm$0.0 & 100.0$\pm$0.0 & 100.0$\pm$0.0 & 100.0$\pm$0.0 & 100.0$\pm$0.0 & 100.0$\pm$0.0 & 100.0$\pm$0.0 \\
      CIFAR-100 & 97.7$\pm$0.0 & 100.0$\pm$0.0 & 98.8$\pm$0.0 & 96.4$\pm$0.0 & 99.9$\pm$0.0 & 100.0$\pm$0.0 & 99.9$\pm$0.0 & 100.0$\pm$0.0 & 100.0$\pm$0.0 & 98.3$\pm$0.0 & 99.9$\pm$0.0 & 100.0$\pm$0.0 & 100.0$\pm$0.0 & 100.0$\pm$0.0 & 100.0$\pm$0.0 \\
      FMNIST-3D & 98.1$\pm$0.0 & 100.0$\pm$0.0 & 98.8$\pm$0.0 & 93.5$\pm$0.0 & 100.0$\pm$0.0 & 100.0$\pm$0.0 & 99.9$\pm$0.0 & 100.0$\pm$0.0 & 100.0$\pm$0.0 & 98.2$\pm$0.0 & 100.0$\pm$0.0 & 100.0$\pm$0.0 & 100.0$\pm$0.0 & 100.0$\pm$0.0 & 100.0$\pm$0.0 \\
      Uniform & 97.3$\pm$0.0 & 100.0$\pm$0.0 & 99.3$\pm$0.0 & 82.8$\pm$0.0 & 100.0$\pm$0.0 & 100.0$\pm$0.0 & 100.0$\pm$0.0 & 100.0$\pm$0.0 & 100.0$\pm$0.0 & 98.7$\pm$0.0 & 100.0$\pm$0.0 & 100.0$\pm$0.0 & 100.0$\pm$0.0 & 100.0$\pm$0.0 & 100.0$\pm$0.0 \\
      Smooth & 98.5$\pm$0.0 & 98.9$\pm$0.0 & 99.4$\pm$0.0 & 98.6$\pm$0.0 & 98.6$\pm$0.0 & 98.0$\pm$0.0 & 96.5$\pm$0.0 & 98.4$\pm$0.0 & 98.7$\pm$0.0 & 99.1$\pm$0.0 & 99.3$\pm$0.0 & 99.5$\pm$0.0 & 98.7$\pm$0.0 & 99.4$\pm$0.0 & 99.5$\pm$0.0 \\

      \midrule

      \textbf{CIFAR-10} & & & & & & \\
      SVHN & 93.3$\pm$0.0 & 96.5$\pm$0.0 & 93.3$\pm$0.0 & 93.3$\pm$0.0 & 92.8$\pm$0.0 & 96.6$\pm$0.0 & 77.6$\pm$0.0 & 94.5$\pm$0.0 & 92.8$\pm$0.0 & 93.3$\pm$0.0 & 95.9$\pm$0.0 & 97.9$\pm$0.0 & 85.2$\pm$0.0 & 97.0$\pm$0.0 & 96.4$\pm$0.0 \\
      LSUN-CR & 99.7$\pm$0.0 & 99.8$\pm$0.0 & 99.7$\pm$0.0 & 99.6$\pm$0.0 & 99.8$\pm$0.0 & 99.9$\pm$0.0 & 99.7$\pm$0.0 & 99.7$\pm$0.0 & 99.7$\pm$0.0 & 99.7$\pm$0.0 & 99.9$\pm$0.0 & 100.0$\pm$0.0 & 99.7$\pm$0.0 & 99.8$\pm$0.0 & 99.8$\pm$0.0 \\
      CIFAR-100 & 89.9$\pm$0.0 & 90.0$\pm$0.0 & 91.3$\pm$0.0 & 86.7$\pm$0.0 & 85.7$\pm$0.0 & 84.2$\pm$0.1 & 82.8$\pm$0.0 & 85.3$\pm$0.0 & 86.9$\pm$0.0 & 90.0$\pm$0.0 & 90.0$\pm$0.0 & 89.1$\pm$0.0 & 82.4$\pm$0.0 & 89.8$\pm$0.0 & 90.0$\pm$0.0 \\
      FMNIST-3D & 95.1$\pm$0.0 & 96.4$\pm$0.0 & 96.1$\pm$0.0 & 94.7$\pm$0.0 & 97.0$\pm$0.0 & 98.5$\pm$0.0 & 92.8$\pm$0.0 & 95.0$\pm$0.0 & 96.1$\pm$0.0 & 95.0$\pm$0.0 & 97.6$\pm$0.0 & 99.0$\pm$0.0 & 93.4$\pm$0.0 & 96.7$\pm$0.0 & 96.3$\pm$0.0 \\
      Uniform & 95.5$\pm$0.0 & 100.0$\pm$0.0 & 99.6$\pm$0.0 & 98.0$\pm$0.0 & 100.0$\pm$0.0 & 100.0$\pm$0.0 & 100.0$\pm$0.0 & 100.0$\pm$0.0 & 100.0$\pm$0.0 & 95.0$\pm$0.0 & 100.0$\pm$0.0 & 100.0$\pm$0.0 & 100.0$\pm$0.0 & 100.0$\pm$0.0 & 100.0$\pm$0.0 \\
      Smooth & 94.4$\pm$0.0 & 97.5$\pm$0.0 & 95.8$\pm$0.0 & 95.7$\pm$0.0 & 95.5$\pm$0.0 & 97.5$\pm$0.0 & 93.9$\pm$0.0 & 95.1$\pm$0.0 & 95.9$\pm$0.0 & 94.7$\pm$0.0 & 97.6$\pm$0.0 & 98.7$\pm$0.0 & 93.9$\pm$0.0 & 98.6$\pm$0.0 & 97.4$\pm$0.0 \\

      \midrule

      \textbf{CIFAR-100} & & & & & & \\
      LSUN-CR & 99.0$\pm$0.0 & 99.4$\pm$0.0 & 99.3$\pm$0.0 & 99.0$\pm$0.0 & 99.5$\pm$0.0 & 99.7$\pm$0.0 & 99.1$\pm$0.0 & 99.3$\pm$0.0 & 99.3$\pm$0.0 & 99.0$\pm$0.0 & 99.5$\pm$0.0 & 99.9$\pm$0.0 & 99.2$\pm$0.0 & 99.4$\pm$0.0 & 99.4$\pm$0.0 \\
      CIFAR-10 & 77.2$\pm$0.0 & 77.0$\pm$0.0 & 79.3$\pm$0.0 & 76.8$\pm$0.0 & 77.1$\pm$0.0 & 73.0$\pm$0.1 & 75.4$\pm$0.0 & 77.4$\pm$0.0 & 77.1$\pm$0.0 & 77.3$\pm$0.0 & 77.1$\pm$0.0 & 68.2$\pm$0.1 & 75.8$\pm$0.0 & 77.2$\pm$0.0 & 77.1$\pm$0.0 \\
      FMNIST-3D & 85.7$\pm$0.0 & 86.0$\pm$0.0 & 88.2$\pm$0.0 & 87.6$\pm$0.0 & 90.9$\pm$0.0 & 96.6$\pm$0.0 & 84.8$\pm$0.0 & 89.3$\pm$0.0 & 87.6$\pm$0.0 & 85.6$\pm$0.0 & 87.1$\pm$0.0 & 96.8$\pm$0.0 & 86.5$\pm$0.0 & 86.3$\pm$0.0 & 85.7$\pm$0.0 \\
      Uniform & 84.6$\pm$0.0 & 100.0$\pm$0.0 & 91.5$\pm$0.0 & 73.3$\pm$0.0 & 100.0$\pm$0.0 & 100.0$\pm$0.0 & 100.0$\pm$0.0 & 100.0$\pm$0.0 & 100.0$\pm$0.0 & 87.1$\pm$0.1 & 100.0$\pm$0.0 & 100.0$\pm$0.0 & 100.0$\pm$0.0 & 99.9$\pm$0.0 & 100.0$\pm$0.0 \\
      Smooth & 75.4$\pm$0.0 & 71.0$\pm$0.0 & 80.3$\pm$0.0 & 67.4$\pm$0.1 & 69.8$\pm$0.1 & 88.1$\pm$0.1 & 81.4$\pm$0.0 & 79.9$\pm$0.1 & 74.9$\pm$0.0 & 78.5$\pm$0.1 & 90.9$\pm$0.0 & 98.8$\pm$0.0 & 78.3$\pm$0.0 & 85.4$\pm$0.1 & 73.9$\pm$0.1 \\

      \bottomrule
    \end{tabular}
  }
\end{table*}

\begin{table*}[t]
  \caption{Confidences in terms of MMC, averaged over five prediction runs. Lower is better for OOD datasets.}
  \label{tab:mmc}

  \centering

  \resizebox{\textwidth}{!}{
    \begin{tabular}{lrrrrrHrrrrrHrrr}
      \toprule

       & & & & \multicolumn{6}{c}{\bf VB} & \multicolumn{6}{c}{\bf LA} \\
       \cmidrule(r){5-10} \cmidrule(l){11-16}
      \textbf{Datasets} & {\bf MAP} & {\bf OE} & {\bf DE} & {\bf Plain} & {\bf NC} & {\bf NC2} & {\bf SL} & {\bf ML} & {\bf OE} & {\bf Plain} & {\bf NC} & {\bf NC2} & {\bf SL} & {\bf ML} & {\bf OE} \\
      \midrule

      \textbf{MNIST} & 99.1 & 99.4 & 99.3 & 98.7 & 98.2 & 98.2 & 99.1 & 98.2 & 98.6 & 99.0 & 99.3 & 99.3 & 99.2 & 99.2 & 99.3 \\
      F-MNIST & 66.3$\pm$0.0 & 22.0$\pm$0.0 & 64.9$\pm$0.0 & 70.4$\pm$0.0 & 6.1$\pm$0.0 & 6.1$\pm$0.0 & 20.4$\pm$0.0 & 21.2$\pm$0.0 & 27.3$\pm$0.0 & 65.2$\pm$0.0 & 7.7$\pm$0.0 & 7.7$\pm$0.0 & 21.0$\pm$0.0 & 20.7$\pm$0.0 & 22.2$\pm$0.0 \\
      E-MNIST & 82.3$\pm$0.0 & 79.3$\pm$0.0 & 80.9$\pm$0.0 & 78.6$\pm$0.0 & 74.0$\pm$0.0 & 73.9$\pm$0.0 & 75.9$\pm$0.0 & 73.0$\pm$0.0 & 73.9$\pm$0.0 & 81.1$\pm$0.0 & 79.6$\pm$0.0 & 79.6$\pm$0.0 & 80.4$\pm$0.0 & 76.5$\pm$0.0 & 78.1$\pm$0.0 \\
      K-MNIST & 73.3$\pm$0.0 & 65.7$\pm$0.0 & 69.7$\pm$0.0 & 67.6$\pm$0.0 & 51.8$\pm$0.0 & 51.8$\pm$0.0 & 61.2$\pm$0.0 & 50.3$\pm$0.0 & 52.4$\pm$0.0 & 71.8$\pm$0.0 & 64.3$\pm$0.0 & 64.3$\pm$0.0 & 66.3$\pm$0.0 & 62.3$\pm$0.0 & 64.3$\pm$0.0 \\
      CIFAR-Gr & 48.0$\pm$0.0 & 10.0$\pm$0.0 & 43.1$\pm$0.0 & 45.2$\pm$0.0 & 0.0$\pm$0.0 & 0.0$\pm$0.0 & 10.0$\pm$0.0 & 10.1$\pm$0.0 & 10.1$\pm$0.0 & 47.3$\pm$0.0 & 0.0$\pm$0.0 & 0.0$\pm$0.0 & 10.1$\pm$0.0 & 10.4$\pm$0.0 & 10.2$\pm$0.0 \\
      Uniform & 96.8$\pm$0.0 & 10.0$\pm$0.0 & 97.4$\pm$0.0 & 97.8$\pm$0.0 & 0.1$\pm$0.0 & 0.1$\pm$0.0 & 10.0$\pm$0.0 & 10.1$\pm$0.0 & 10.1$\pm$0.0 & 96.5$\pm$0.0 & 0.0$\pm$0.0 & 0.0$\pm$0.0 & 10.2$\pm$0.0 & 10.3$\pm$0.0 & 10.2$\pm$0.0 \\
      Smooth & 12.9$\pm$0.0 & 10.1$\pm$0.0 & 12.6$\pm$0.0 & 12.7$\pm$0.0 & 0.3$\pm$0.0 & 0.3$\pm$0.0 & 21.6$\pm$0.0 & 10.2$\pm$0.0 & 10.2$\pm$0.0 & 12.8$\pm$0.0 & 0.5$\pm$0.0 & 0.5$\pm$0.0 & 20.5$\pm$0.0 & 10.1$\pm$0.0 & 10.1$\pm$0.0 \\

      \midrule

      \textbf{F-MNIST} & 96.1 & 96.0 & 94.8 & 93.6 & 91.0 & 91.0 & 95.0 & 91.2 & 92.9 & 95.7 & 94.9 & 94.9 & 93.5 & 94.5 & 94.7 \\
      MNIST & 82.8$\pm$0.0 & 60.1$\pm$0.0 & 74.3$\pm$0.0 & 70.9$\pm$0.0 & 50.0$\pm$0.0 & 50.0$\pm$0.0 & 32.9$\pm$0.0 & 57.2$\pm$0.0 & 63.7$\pm$0.0 & 80.9$\pm$0.0 & 33.8$\pm$0.1 & 33.8$\pm$0.1 & 33.9$\pm$0.0 & 55.9$\pm$0.1 & 57.9$\pm$0.0 \\
      E-MNIST & 82.5$\pm$0.0 & 45.0$\pm$0.0 & 70.0$\pm$0.0 & 71.5$\pm$0.0 & 25.0$\pm$0.0 & 25.0$\pm$0.0 & 25.5$\pm$0.0 & 44.4$\pm$0.0 & 50.7$\pm$0.0 & 80.6$\pm$0.0 & 9.4$\pm$0.0 & 9.4$\pm$0.0 & 24.3$\pm$0.0 & 36.2$\pm$0.0 & 44.7$\pm$0.0 \\
      K-MNIST & 82.5$\pm$0.0 & 60.1$\pm$0.0 & 64.0$\pm$0.0 & 68.4$\pm$0.0 & 37.0$\pm$0.0 & 37.1$\pm$0.0 & 33.5$\pm$0.0 & 44.8$\pm$0.0 & 48.0$\pm$0.0 & 80.0$\pm$0.0 & 28.1$\pm$0.0 & 28.1$\pm$0.0 & 32.0$\pm$0.0 & 52.6$\pm$0.0 & 57.3$\pm$0.0 \\
      CIFAR-Gr & 89.1$\pm$0.0 & 10.0$\pm$0.0 & 84.6$\pm$0.0 & 73.1$\pm$0.0 & 0.0$\pm$0.0 & 0.0$\pm$0.0 & 10.0$\pm$0.0 & 10.1$\pm$0.0 & 10.2$\pm$0.0 & 88.6$\pm$0.0 & 0.0$\pm$0.0 & 0.0$\pm$0.0 & 10.8$\pm$0.0 & 10.3$\pm$0.0 & 10.3$\pm$0.0 \\
      Uniform & 85.5$\pm$0.0 & 13.2$\pm$0.0 & 82.3$\pm$0.0 & 79.4$\pm$0.0 & 0.6$\pm$0.0 & 0.6$\pm$0.0 & 10.2$\pm$0.0 & 10.6$\pm$0.0 & 11.1$\pm$0.0 & 84.2$\pm$0.0 & 0.0$\pm$0.0 & 0.0$\pm$0.0 & 10.9$\pm$0.0 & 12.2$\pm$0.0 & 15.0$\pm$0.0 \\
      Smooth & 48.0$\pm$0.0 & 10.5$\pm$0.0 & 44.0$\pm$0.0 & 42.0$\pm$0.0 & 1.0$\pm$0.0 & 1.0$\pm$0.0 & 11.0$\pm$0.0 & 11.9$\pm$0.0 & 10.7$\pm$0.0 & 47.4$\pm$0.0 & 0.3$\pm$0.0 & 0.3$\pm$0.0 & 11.6$\pm$0.0 & 10.4$\pm$0.0 & 10.8$\pm$0.0 \\

      \midrule

      \textbf{SVHN} & 98.6 & 98.6 & 98.1 & 97.7 & 97.3 & 97.3 & 98.4 & 97.1 & 97.6 & 97.9 & 98.0 & 98.0 & 97.9 & 97.8 & 97.3 \\
      CIFAR-10 & 69.0$\pm$0.0 & 11.6$\pm$0.0 & 57.3$\pm$0.0 & 61.3$\pm$0.0 & 3.7$\pm$0.0 & 3.7$\pm$0.0 & 15.4$\pm$0.0 & 11.1$\pm$0.0 & 11.7$\pm$0.0 & 60.7$\pm$0.0 & 3.6$\pm$0.0 & 3.5$\pm$0.0 & 12.4$\pm$0.0 & 12.1$\pm$0.0 & 12.5$\pm$0.0 \\
      LSUN-CR & 69.8$\pm$0.0 & 10.2$\pm$0.0 & 57.7$\pm$0.0 & 63.9$\pm$0.0 & 0.1$\pm$0.0 & 0.1$\pm$0.0 & 10.8$\pm$0.0 & 10.2$\pm$0.0 & 10.2$\pm$0.0 & 61.5$\pm$0.1 & 0.0$\pm$0.0 & 0.0$\pm$0.0 & 11.0$\pm$0.0 & 10.3$\pm$0.0 & 10.7$\pm$0.0 \\
      CIFAR-100 & 70.3$\pm$0.0 & 12.4$\pm$0.0 & 58.8$\pm$0.0 & 63.5$\pm$0.0 & 4.3$\pm$0.0 & 4.3$\pm$0.0 & 17.3$\pm$0.0 & 11.6$\pm$0.0 & 12.3$\pm$0.0 & 62.3$\pm$0.0 & 4.1$\pm$0.0 & 4.1$\pm$0.0 & 13.4$\pm$0.0 & 12.7$\pm$0.0 & 13.4$\pm$0.0 \\
      FMNIST-3D & 73.6$\pm$0.0 & 11.0$\pm$0.0 & 62.5$\pm$0.0 & 67.8$\pm$0.0 & 1.5$\pm$0.0 & 1.5$\pm$0.0 & 17.6$\pm$0.0 & 10.8$\pm$0.0 & 11.5$\pm$0.0 & 68.0$\pm$0.0 & 1.8$\pm$0.0 & 1.8$\pm$0.0 & 15.5$\pm$0.0 & 10.9$\pm$0.0 & 12.0$\pm$0.0 \\
      Uniform & 77.5$\pm$0.0 & 10.3$\pm$0.0 & 57.1$\pm$0.0 & 82.2$\pm$0.0 & 0.1$\pm$0.0 & 0.1$\pm$0.0 & 10.0$\pm$0.0 & 10.1$\pm$0.0 & 10.2$\pm$0.0 & 65.7$\pm$0.1 & 0.0$\pm$0.0 & 0.0$\pm$0.0 & 10.4$\pm$0.0 & 10.3$\pm$0.0 & 10.7$\pm$0.0 \\
      Smooth & 68.8$\pm$0.0 & 52.1$\pm$0.0 & 51.3$\pm$0.0 & 55.5$\pm$0.0 & 42.3$\pm$0.0 & 42.3$\pm$0.0 & 62.7$\pm$0.0 & 45.6$\pm$0.0 & 44.0$\pm$0.0 & 58.3$\pm$0.1 & 43.9$\pm$0.0 & 43.8$\pm$0.0 & 44.9$\pm$0.0 & 44.6$\pm$0.1 & 40.8$\pm$0.0 \\

      \midrule

      \textbf{CIFAR-10} & 96.7 & 96.9 & 95.9 & 95.7 & 94.5 & 94.5 & 96.0 & 95.1 & 95.5 & 96.4 & 94.9 & 94.9 & 95.8 & 96.1 & 96.1 \\
      SVHN & 65.2$\pm$0.0 & 45.7$\pm$0.0 & 60.8$\pm$0.0 & 59.5$\pm$0.0 & 49.7$\pm$0.0 & 49.8$\pm$0.0 & 73.3$\pm$0.0 & 45.0$\pm$0.0 & 55.8$\pm$0.0 & 63.5$\pm$0.0 & 36.8$\pm$0.1 & 36.9$\pm$0.0 & 64.7$\pm$0.0 & 33.0$\pm$0.0 & 44.0$\pm$0.0 \\
      LSUN-CR & 73.9$\pm$0.0 & 58.7$\pm$0.0 & 65.2$\pm$0.0 & 68.2$\pm$0.0 & 41.1$\pm$0.1 & 41.1$\pm$0.1 & 64.3$\pm$0.0 & 56.7$\pm$0.0 & 57.1$\pm$0.0 & 71.6$\pm$0.0 & 29.2$\pm$0.2 & 28.6$\pm$0.2 & 61.7$\pm$0.4 & 53.6$\pm$0.3 & 55.2$\pm$0.3 \\
      CIFAR-100 & 77.2$\pm$0.0 & 76.1$\pm$0.0 & 69.8$\pm$0.0 & 72.4$\pm$0.0 & 67.7$\pm$0.0 & 67.7$\pm$0.0 & 75.9$\pm$0.0 & 69.8$\pm$0.0 & 70.5$\pm$0.0 & 75.4$\pm$0.0 & 67.3$\pm$0.0 & 67.3$\pm$0.0 & 74.3$\pm$0.0 & 73.2$\pm$0.0 & 72.6$\pm$0.0 \\
      FMNIST-3D & 67.7$\pm$0.0 & 55.9$\pm$0.0 & 58.5$\pm$0.0 & 60.4$\pm$0.0 & 32.7$\pm$0.0 & 32.7$\pm$0.0 & 64.3$\pm$0.0 & 54.5$\pm$0.0 & 52.3$\pm$0.0 & 66.0$\pm$0.0 & 27.4$\pm$0.0 & 27.3$\pm$0.0 & 57.6$\pm$0.0 & 48.5$\pm$0.0 & 53.1$\pm$0.1 \\
      Uniform & 82.1$\pm$0.0 & 10.3$\pm$0.0 & 40.7$\pm$0.0 & 52.2$\pm$0.0 & 0.1$\pm$0.0 & 0.1$\pm$0.0 & 10.2$\pm$0.0 & 10.7$\pm$0.0 & 10.1$\pm$0.0 & 81.9$\pm$0.1 & 0.0$\pm$0.0 & 0.0$\pm$0.0 & 10.2$\pm$0.0 & 10.3$\pm$0.0 & 10.3$\pm$0.0 \\
      Smooth & 64.3$\pm$0.0 & 45.2$\pm$0.0 & 57.2$\pm$0.0 & 54.0$\pm$0.0 & 51.4$\pm$0.0 & 51.5$\pm$0.0 & 53.3$\pm$0.0 & 53.0$\pm$0.0 & 49.3$\pm$0.0 & 62.5$\pm$0.0 & 32.6$\pm$0.0 & 32.7$\pm$0.1 & 59.5$\pm$0.1 & 26.4$\pm$0.0 & 43.5$\pm$0.0 \\

      \midrule

      \textbf{CIFAR-100} & 84.7 & 85.1 & 80.9 & 69.0 & 66.2 & 66.2 & 80.1 & 67.1 & 67.1 & 81.6 & 80.3 & 80.3 & 77.0 & 79.1 & 79.4 \\
      SVHN & 52.8$\pm$0.0 & 43.9$\pm$0.0 & 43.8$\pm$0.0 & 27.3$\pm$0.0 & 31.0$\pm$0.0 & 30.9$\pm$0.0 & 46.2$\pm$0.0 & 24.9$\pm$0.0 & 26.9$\pm$0.0 & 46.9$\pm$0.0 & 33.7$\pm$0.1 & 33.8$\pm$0.0 & 42.1$\pm$0.0 & 34.8$\pm$0.0 & 37.9$\pm$0.1 \\
      LSUN-CR & 62.7$\pm$0.0 & 51.3$\pm$0.0 & 48.8$\pm$0.0 & 33.3$\pm$0.1 & 21.1$\pm$0.0 & 21.2$\pm$0.0 & 49.7$\pm$0.0 & 26.5$\pm$0.0 & 26.4$\pm$0.1 & 57.0$\pm$0.1 & 36.4$\pm$0.1 & 36.3$\pm$0.1 & 43.5$\pm$0.5 & 40.9$\pm$0.1 & 42.6$\pm$0.3 \\
      CIFAR-10 & 62.9$\pm$0.0 & 63.7$\pm$0.0 & 53.4$\pm$0.0 & 39.0$\pm$0.0 & 36.9$\pm$0.0 & 36.9$\pm$0.0 & 55.3$\pm$0.0 & 37.2$\pm$0.0 & 37.4$\pm$0.0 & 57.3$\pm$0.0 & 55.7$\pm$0.0 & 55.6$\pm$0.0 & 50.8$\pm$0.0 & 53.8$\pm$0.0 & 54.0$\pm$0.0 \\
      FMNIST-3D & 51.8$\pm$0.0 & 48.2$\pm$0.0 & 42.5$\pm$0.0 & 24.8$\pm$0.0 & 15.9$\pm$0.0 & 15.9$\pm$0.0 & 44.5$\pm$0.0 & 20.6$\pm$0.0 & 22.6$\pm$0.0 & 47.1$\pm$0.0 & 40.7$\pm$0.0 & 40.8$\pm$0.0 & 38.4$\pm$0.0 & 40.3$\pm$0.0 & 41.4$\pm$0.0 \\
      Uniform & 64.2$\pm$0.0 & 1.4$\pm$0.0 & 45.0$\pm$0.0 & 59.4$\pm$0.0 & 0.0$\pm$0.0 & 0.0$\pm$0.0 & 2.2$\pm$0.0 & 1.2$\pm$0.0 & 1.2$\pm$0.0 & 54.4$\pm$0.1 & 0.0$\pm$0.0 & 0.0$\pm$0.0 & 1.5$\pm$0.0 & 4.4$\pm$0.1 & 1.5$\pm$0.0 \\
      Smooth & 61.7$\pm$0.0 & 58.9$\pm$0.0 & 47.3$\pm$0.0 & 49.3$\pm$0.0 & 38.4$\pm$0.0 & 38.5$\pm$0.0 & 50.6$\pm$0.0 & 28.7$\pm$0.0 & 34.8$\pm$0.0 & 54.8$\pm$0.1 & 30.4$\pm$0.1 & 30.5$\pm$0.1 & 49.1$\pm$0.0 & 40.2$\pm$0.1 & 51.5$\pm$0.1 \\

      \bottomrule
    \end{tabular}
  }
\end{table*}

\begin{table*}[t]
  \caption{Test accuracy (\(\uparrow\)) / ECE (\(\downarrow\)) of models trained with random noises \citep{hein2019relu} as \(\Dout\), averaged over five prediction runs.}
  \label{tab:acc_ece_smooth}

  \centering
  \small
  \renewcommand{\tabcolsep}{10pt}

  \resizebox{\textwidth}{!}{
    \begin{tabular}{lllllll}
      \toprule
      \textbf{\footnotesize} & {\footnotesize\bf MNIST} & {\footnotesize\bf F-MNIST} & {\footnotesize\bf SVHN} & {\footnotesize\bf CIFAR-10} & {\footnotesize\bf CIFAR-100} \\
      \midrule

      MAP & 99.4$\pm$0.0\,/\,6.4$\pm$0.0 & 92.4$\pm$0.0\,/\,13.9$\pm$0.0 & 97.4$\pm$0.0\,/\,8.9$\pm$0.0 & 94.8$\pm$0.0\,/\,10.0$\pm$0.0 & 76.7$\pm$0.0\,/\,14.3$\pm$0.0 \\
      DE & 99.5$\pm$0.0\,/\,8.6$\pm$0.0 & 93.6$\pm$0.0\,/\,3.6$\pm$0.0 & 97.6$\pm$0.0\,/\,3.5$\pm$0.0 & 95.7$\pm$0.0\,/\,4.5$\pm$0.0 & 80.0$\pm$0.0\,/\,1.9$\pm$0.0 \\
      OE & 99.6$\pm$0.0\,/\,6.4$\pm$0.0 & 92.6$\pm$0.0\,/\,12.7$\pm$0.0 & 97.5$\pm$0.0\,/\,8.9$\pm$0.0 & 94.7$\pm$0.0\,/\,11.5$\pm$0.0 & 76.5$\pm$0.0\,/\,16.1$\pm$0.0 \\

      \midrule

      VB & 99.5$\pm$0.0\,/\,11.2$\pm$0.3 & 92.4$\pm$0.0\,/\,3.7$\pm$0.2 & 97.5$\pm$0.0\,/\,5.7$\pm$0.2 & 94.9$\pm$0.0\,/\,5.8$\pm$0.2 & 75.4$\pm$0.0\,/\,8.3$\pm$0.0 \\
      +NC & 99.4$\pm$0.0\,/\,10.6$\pm$0.1 & 92.3$\pm$0.0\,/\,3.0$\pm$0.1 & 97.4$\pm$0.0\,/\,4.2$\pm$0.2 & 94.9$\pm$0.0\,/\,5.1$\pm$0.1 & 74.0$\pm$0.1\,/\,8.8$\pm$0.1 \\
      +SL & 99.6$\pm$0.0\,/\,12.4$\pm$0.1 & 93.2$\pm$0.0\,/\,12.4$\pm$0.1 & 97.3$\pm$0.0\,/\,12.8$\pm$0.0 & 91.5$\pm$0.0\,/\,18.5$\pm$0.1 & 1.0$\pm$0.1\,/\,0.1$\pm$0.0 \\
      +ML & 99.4$\pm$0.0\,/\,11.6$\pm$0.2 & 92.1$\pm$0.0\,/\,2.4$\pm$0.1 & 97.6$\pm$0.0\,/\,3.3$\pm$0.1 & 95.1$\pm$0.0\,/\,3.6$\pm$0.2 & 75.2$\pm$0.0\,/\,9.8$\pm$0.0 \\
      +OE & 99.5$\pm$0.0\,/\,10.2$\pm$0.2 & 92.5$\pm$0.0\,/\,3.2$\pm$0.1 & 97.5$\pm$0.0\,/\,5.0$\pm$0.1 & 94.9$\pm$0.0\,/\,6.8$\pm$0.2 & 74.1$\pm$0.0\,/\,8.0$\pm$0.0 \\

      \midrule

      LA & 99.4$\pm$0.0\,/\,7.6$\pm$0.1 & 92.5$\pm$0.0\,/\,11.3$\pm$0.2 & 97.4$\pm$0.0\,/\,3.3$\pm$0.3 & 94.8$\pm$0.0\,/\,7.5$\pm$0.3 & 76.6$\pm$0.1\,/\,8.3$\pm$0.1 \\
      +NC & 99.3$\pm$0.0\,/\,10.1$\pm$0.8 & 92.5$\pm$0.0\,/\,2.8$\pm$0.1 & 96.3$\pm$0.0\,/\,5.0$\pm$0.1 & 94.9$\pm$0.0\,/\,8.3$\pm$0.3 & 75.9$\pm$0.1\,/\,3.8$\pm$0.1 \\
      +SL & 99.6$\pm$0.0\,/\,11.1$\pm$0.4 & 93.0$\pm$0.0\,/\,9.1$\pm$0.1 & 18.8$\pm$0.0\,/\,13.5$\pm$0.1 & 91.5$\pm$0.0\,/\,16.0$\pm$0.2 & 72.2$\pm$0.0\,/\,4.0$\pm$0.1 \\
      +ML & 99.5$\pm$0.0\,/\,5.5$\pm$0.2 & 92.3$\pm$0.0\,/\,9.3$\pm$0.1 & 97.5$\pm$0.0\,/\,7.4$\pm$0.3 & 94.8$\pm$0.0\,/\,7.2$\pm$0.3 & 76.8$\pm$0.1\,/\,3.3$\pm$0.2 \\
      +OE & 99.6$\pm$0.0\,/\,6.6$\pm$0.3 & 92.3$\pm$0.0\,/\,2.0$\pm$0.1 & 97.5$\pm$0.0\,/\,3.8$\pm$0.2 & 94.6$\pm$0.0\,/\,7.2$\pm$0.3 & 76.3$\pm$0.0\,/\,8.6$\pm$0.1 \\

      \bottomrule
    \end{tabular}
  }
\end{table*}

\begin{table*}[ht]
  \caption{OOD data detection under models trained with random noises \citep{hein2019relu} as $\Dout$. Values are FPR95, averaged over five prediction runs---lower is better.}
  \label{tab:fpr95_noise_more}

  \centering

  \resizebox{\textwidth}{!}{
    \begin{tabular}{lrrrrrHrrrrrHrrr}
      \toprule

      & & & & \multicolumn{6}{c}{\bf VB} & \multicolumn{6}{c}{\bf LA} \\
      \cmidrule(r){5-10} \cmidrule(l){11-16}
      \textbf{Datasets} & {\bf MAP} & {\bf OE} & {\bf DE} & {\bf Plain} & {\bf NC} & {\bf NC2} & {\bf SL} & {\bf ML} & {\bf OE} & {\bf Plain} & {\bf NC} & {\bf NC2} & {\bf SL} & {\bf ML} & {\bf OE} \\
      \midrule

      \textbf{MNIST} & & & & & & \\
      F-MNIST & 11.8$\pm$0.0 & 6.8$\pm$0.0 & 5.3$\pm$0.0 & 12.5$\pm$0.1 & 6.5$\pm$0.0 & 6.5$\pm$0.2 & 0.7$\pm$0.0 & 11.9$\pm$0.1 & 10.3$\pm$0.1 & 12.0$\pm$0.0 & 8.2$\pm$0.0 & 4.3$\pm$0.0 & 0.0$\pm$0.0 & 6.3$\pm$0.0 & 6.8$\pm$0.0 \\
      E-MNIST & 35.6$\pm$0.0 & 30.7$\pm$0.0 & 30.4$\pm$0.0 & 34.5$\pm$0.1 & 35.1$\pm$0.1 & 42.4$\pm$0.6 & 17.9$\pm$0.0 & 37.3$\pm$0.1 & 34.3$\pm$0.1 & 35.8$\pm$0.1 & 34.2$\pm$0.0 & 39.2$\pm$0.1 & 15.3$\pm$0.1 & 31.0$\pm$0.0 & 30.7$\pm$0.0 \\
      K-MNIST & 14.4$\pm$0.0 & 7.8$\pm$0.0 & 7.7$\pm$0.0 & 14.0$\pm$0.1 & 14.5$\pm$0.1 & 17.1$\pm$0.4 & 1.1$\pm$0.0 & 15.8$\pm$0.1 & 14.0$\pm$0.1 & 14.5$\pm$0.1 & 10.6$\pm$0.0 & 9.5$\pm$0.0 & 0.7$\pm$0.0 & 8.5$\pm$0.0 & 7.8$\pm$0.0 \\
      CIFAR-Gr & 0.2$\pm$0.0 & 0.0$\pm$0.0 & 0.0$\pm$0.0 & 0.2$\pm$0.0 & 0.0$\pm$0.0 & 0.0$\pm$0.0 & 0.0$\pm$0.0 & 0.0$\pm$0.0 & 0.1$\pm$0.0 & 0.2$\pm$0.0 & 0.0$\pm$0.0 & 0.0$\pm$0.0 & 0.0$\pm$0.0 & 0.0$\pm$0.0 & 0.0$\pm$0.0 \\
      Uniform & 44.3$\pm$0.0 & 0.7$\pm$0.0 & 19.8$\pm$0.0 & 93.1$\pm$0.2 & 0.0$\pm$0.0 & 0.0$\pm$0.0 & 0.0$\pm$0.0 & 1.9$\pm$0.1 & 1.8$\pm$0.0 & 54.2$\pm$0.4 & 0.7$\pm$0.0 & 0.0$\pm$0.0 & 0.0$\pm$0.0 & 0.6$\pm$0.0 & 0.8$\pm$0.0 \\

      \midrule

      \textbf{F-MNIST} & & & & & & \\
      MNIST & 73.5$\pm$0.0 & 62.2$\pm$0.0 & 65.8$\pm$0.0 & 66.8$\pm$0.1 & 62.2$\pm$0.0 & 34.6$\pm$0.2 & 30.4$\pm$0.1 & 59.1$\pm$0.1 & 60.4$\pm$0.1 & 72.2$\pm$0.2 & 60.8$\pm$0.2 & 13.8$\pm$0.2 & 24.3$\pm$0.3 & 55.4$\pm$0.2 & 57.4$\pm$0.3 \\
      E-MNIST & 73.6$\pm$0.0 & 50.2$\pm$0.0 & 58.6$\pm$0.0 & 68.1$\pm$0.1 & 43.9$\pm$0.0 & 13.0$\pm$0.1 & 25.7$\pm$0.0 & 54.7$\pm$0.1 & 54.6$\pm$0.1 & 72.2$\pm$0.2 & 44.2$\pm$0.1 & 3.3$\pm$0.1 & 22.5$\pm$0.2 & 39.6$\pm$0.2 & 48.3$\pm$0.3 \\
      K-MNIST & 73.7$\pm$0.0 & 47.4$\pm$0.0 & 47.2$\pm$0.0 & 62.6$\pm$0.1 & 31.4$\pm$0.1 & 8.2$\pm$0.0 & 19.1$\pm$0.0 & 35.6$\pm$0.1 & 38.1$\pm$0.1 & 71.5$\pm$0.2 & 33.9$\pm$0.2 & 2.9$\pm$0.1 & 20.9$\pm$0.2 & 31.7$\pm$0.2 & 43.0$\pm$0.4 \\
      CIFAR-Gr & 87.2$\pm$0.0 & 0.5$\pm$0.0 & 86.6$\pm$0.0 & 75.3$\pm$0.0 & 0.1$\pm$0.0 & 0.0$\pm$0.0 & 0.2$\pm$0.0 & 0.8$\pm$0.0 & 1.1$\pm$0.0 & 87.7$\pm$0.1 & 0.7$\pm$0.0 & 0.0$\pm$0.0 & 0.2$\pm$0.0 & 0.7$\pm$0.0 & 1.0$\pm$0.0 \\
      Uniform & 81.3$\pm$0.0 & 26.0$\pm$0.0 & 86.3$\pm$0.0 & 87.3$\pm$0.1 & 47.1$\pm$0.2 & 0.6$\pm$0.0 & 0.0$\pm$0.0 & 0.1$\pm$0.0 & 4.9$\pm$0.0 & 81.0$\pm$0.2 & 43.4$\pm$0.7 & 0.0$\pm$0.0 & 0.0$\pm$0.0 & 22.0$\pm$0.2 & 38.1$\pm$0.9 \\

      \midrule

      \textbf{SVHN} & & & & & & \\
      CIFAR-10 & 18.9$\pm$0.0 & 13.8$\pm$0.0 & 9.5$\pm$0.0 & 15.0$\pm$0.0 & 13.0$\pm$0.1 & 14.5$\pm$0.0 & 16.5$\pm$0.0 & 8.4$\pm$0.0 & 11.5$\pm$0.0 & 15.4$\pm$0.1 & 8.4$\pm$0.0 & 18.7$\pm$0.1 & 94.8$\pm$1.6 & 14.9$\pm$0.1 & 11.4$\pm$0.1 \\
      LSUN-CR & 19.7$\pm$0.0 & 9.0$\pm$0.0 & 8.3$\pm$0.0 & 17.2$\pm$0.2 & 10.5$\pm$0.1 & 8.8$\pm$0.2 & 8.8$\pm$0.1 & 5.4$\pm$0.1 & 9.1$\pm$0.1 & 15.5$\pm$0.1 & 8.3$\pm$0.3 & 3.5$\pm$0.2 & 95.4$\pm$4.0 & 12.6$\pm$0.2 & 8.2$\pm$0.1 \\
      CIFAR-100 & 21.8$\pm$0.0 & 15.6$\pm$0.0 & 11.6$\pm$0.0 & 18.1$\pm$0.0 & 14.8$\pm$0.1 & 16.4$\pm$0.1 & 17.9$\pm$0.0 & 10.2$\pm$0.0 & 12.4$\pm$0.0 & 17.6$\pm$0.1 & 11.7$\pm$0.0 & 21.4$\pm$0.0 & 93.9$\pm$0.7 & 16.6$\pm$0.1 & 13.4$\pm$0.1 \\
      FMNIST-3D & 26.7$\pm$0.0 & 29.8$\pm$0.0 & 17.5$\pm$0.0 & 24.5$\pm$0.1 & 31.1$\pm$0.0 & 34.8$\pm$0.1 & 30.4$\pm$0.0 & 30.0$\pm$0.1 & 25.3$\pm$0.0 & 27.2$\pm$0.1 & 34.6$\pm$0.1 & 70.5$\pm$0.1 & 95.1$\pm$0.8 & 23.3$\pm$0.1 & 27.7$\pm$0.1 \\
      Uniform & 30.0$\pm$0.0 & 0.0$\pm$0.0 & 6.4$\pm$0.0 & 48.2$\pm$0.0 & 0.0$\pm$0.0 & 0.0$\pm$0.0 & 0.0$\pm$0.0 & 0.0$\pm$0.0 & 0.0$\pm$0.0 & 17.0$\pm$0.1 & 0.0$\pm$0.0 & 0.0$\pm$0.0 & 90.2$\pm$0.6 & 19.0$\pm$0.1 & 0.0$\pm$0.0 \\

      \midrule

      \textbf{CIFAR-10} & & & & & & \\
      SVHN & 34.5$\pm$0.0 & 7.3$\pm$0.0 & 33.9$\pm$0.0 & 33.5$\pm$0.0 & 11.1$\pm$0.0 & 3.8$\pm$0.0 & 20.6$\pm$0.0 & 11.0$\pm$0.0 & 9.7$\pm$0.0 & 35.5$\pm$0.1 & 6.6$\pm$0.0 & 0.8$\pm$0.0 & 26.0$\pm$0.3 & 7.5$\pm$0.0 & 8.3$\pm$0.1 \\
      LSUN-CR & 53.3$\pm$0.0 & 49.0$\pm$0.0 & 44.0$\pm$0.0 & 49.4$\pm$0.4 & 46.7$\pm$0.1 & 35.0$\pm$0.5 & 61.7$\pm$0.0 & 45.7$\pm$0.2 & 47.6$\pm$0.3 & 53.8$\pm$0.6 & 48.9$\pm$0.5 & 27.8$\pm$0.3 & 54.9$\pm$0.4 & 51.9$\pm$0.3 & 48.3$\pm$0.5 \\
      CIFAR-100 & 61.2$\pm$0.0 & 58.2$\pm$0.0 & 52.5$\pm$0.0 & 58.4$\pm$0.1 & 57.7$\pm$0.1 & 49.6$\pm$0.1 & 71.1$\pm$0.0 & 56.3$\pm$0.1 & 56.6$\pm$0.1 & 61.4$\pm$0.1 & 59.3$\pm$0.2 & 49.5$\pm$0.2 & 70.7$\pm$0.3 & 57.6$\pm$0.1 & 59.0$\pm$0.3 \\
      FMNIST-3D & 42.4$\pm$0.0 & 44.9$\pm$0.0 & 30.7$\pm$0.0 & 37.4$\pm$0.0 & 39.5$\pm$0.1 & 24.3$\pm$0.1 & 62.7$\pm$0.0 & 43.3$\pm$0.1 & 44.0$\pm$0.1 & 43.2$\pm$0.2 & 40.2$\pm$0.1 & 14.1$\pm$0.1 & 57.8$\pm$0.4 & 40.8$\pm$0.1 & 46.4$\pm$0.3 \\
      Uniform & 87.7$\pm$0.0 & 26.7$\pm$0.0 & 0.0$\pm$0.0 & 13.8$\pm$0.1 & 100.0$\pm$0.0 & 100.0$\pm$0.0 & 57.3$\pm$0.1 & 98.0$\pm$0.0 & 100.0$\pm$0.0 & 92.8$\pm$0.1 & 3.5$\pm$0.1 & 0.0$\pm$0.0 & 17.7$\pm$0.4 & 12.9$\pm$0.3 & 49.8$\pm$0.9 \\

      \midrule

      \textbf{CIFAR-100} & & & & & & \\
      LSUN-CR & 82.0$\pm$0.0 & 79.7$\pm$0.0 & 75.3$\pm$0.0 & 73.8$\pm$0.2 & 80.9$\pm$0.3 & 83.9$\pm$0.4 & 91.5$\pm$8.5 & 77.9$\pm$0.1 & 71.1$\pm$0.4 & 82.8$\pm$0.5 & 82.0$\pm$0.8 & 85.3$\pm$0.5 & 78.5$\pm$0.8 & 72.8$\pm$0.9 & 79.7$\pm$0.8 \\
      CIFAR-10 & 79.8$\pm$0.0 & 80.5$\pm$0.0 & 76.4$\pm$0.0 & 78.2$\pm$0.1 & 81.3$\pm$0.0 & 86.9$\pm$0.2 & 93.9$\pm$1.1 & 80.0$\pm$0.1 & 81.4$\pm$0.1 & 79.5$\pm$0.1 & 80.6$\pm$0.2 & 88.9$\pm$0.1 & 82.1$\pm$0.1 & 78.8$\pm$0.2 & 80.2$\pm$0.2 \\
      FMNIST-3D & 65.8$\pm$0.0 & 66.9$\pm$0.0 & 61.8$\pm$0.0 & 57.1$\pm$0.1 & 69.3$\pm$0.1 & 42.1$\pm$0.4 & 93.6$\pm$1.1 & 63.1$\pm$0.1 & 61.9$\pm$0.1 & 66.1$\pm$0.1 & 71.2$\pm$0.2 & 60.8$\pm$0.3 & 82.0$\pm$0.1 & 64.4$\pm$0.4 & 67.9$\pm$0.2 \\
      Uniform & 97.6$\pm$0.0 & 73.3$\pm$0.0 & 94.3$\pm$0.0 & 100.0$\pm$0.0 & 99.7$\pm$0.0 & 49.5$\pm$1.7 & 95.4$\pm$1.4 & 99.5$\pm$0.0 & 99.8$\pm$0.0 & 98.8$\pm$0.1 & 88.4$\pm$0.5 & 100.0$\pm$0.0 & 100.0$\pm$0.0 & 88.9$\pm$0.5 & 54.0$\pm$0.6 \\

      \bottomrule
    \end{tabular}
  }
\end{table*}

\begin{table*}[ht]
  \caption{Accuracy and ECE on text classification tasks, averaged over five prediction runs.}
  \label{tab:acc_ece_nlp}

  \centering
  \small
  \renewcommand{\tabcolsep}{10pt}

  \begin{tabular}{lll}
    \toprule
    \textbf{Methods} & {\bf SST} & {\bf TREC} \\
    \midrule

    MAP & 78.1$\pm$0.0\,/\,20.8$\pm$0.0 & 76.0$\pm$0.0\,/\,17.2$\pm$0.0 \\
    DE & 82.9$\pm$0.0\,/\,2.5$\pm$0.0 & 80.6$\pm$0.0\,/\,10.6$\pm$0.0 \\
    OE & 78.6$\pm$0.0\,/\,13.0$\pm$0.0 & 68.8$\pm$0.0\,/\,9.4$\pm$0.0 \\

    \midrule

    LA & 78.0$\pm$0.0\,/\,21.0$\pm$0.4 & 75.9$\pm$0.1\,/\,17.3$\pm$0.3 \\
    +NC & 78.0$\pm$0.0\,/\,17.9$\pm$0.1 & 43.4$\pm$0.0\,/\,18.6$\pm$0.2 \\
    +DL & 69.2$\pm$0.2\,/\,17.5$\pm$0.7 & 45.0$\pm$0.2\,/\,10.4$\pm$0.8 \\
    +ML & 48.9$\pm$0.3\,/\,11.4$\pm$0.2 & 55.8$\pm$0.1\,/\,11.5$\pm$0.3 \\
    +OE & 78.5$\pm$0.0\,/\,12.8$\pm$0.4 & 68.1$\pm$0.1\,/\,8.4$\pm$0.7 \\

    \bottomrule
  \end{tabular}
\end{table*}

\begin{table*}[ht]
  \caption{OOD data detection on text classification tasks. Values are FPR95, averaged over five prediction runs---lower is better.}
  \label{tab:fpr95_nlp_detailed}

  \centering
  \small
  \renewcommand{\tabcolsep}{10pt}

  \begin{tabular}{lrrrrrHrrr}
    \toprule

    & & & & \multicolumn{6}{c}{\bf LA} \\
    \cmidrule(r){5-10}
    \textbf{Datasets} & {\bf MAP} & {\bf OE} & {\bf DE} & {\bf Plain} & {\bf NC} & {\bf NC2} & {\bf SL} & {\bf ML} & {\bf OE} \\
    \midrule

    \textbf{SST} & & & & & & \\
    SNLI & 100.0$\pm$0.0 & 0.0$\pm$0.0 & 100.0$\pm$0.0 & 100.0$\pm$0.0 & 0.0$\pm$0.0 & 0.0$\pm$0.0 & 97.0$\pm$0.3 & 89.6$\pm$0.7 & 0.0$\pm$0.0 \\
    Multi30k & 100.0$\pm$0.0 & 0.0$\pm$0.0 & 100.0$\pm$0.0 & 100.0$\pm$0.0 & 0.0$\pm$0.0 & 0.0$\pm$0.0 & 99.5$\pm$0.0 & 83.5$\pm$1.5 & 0.0$\pm$0.0 \\
    WMT16 & 100.0$\pm$0.0 & 0.0$\pm$0.0 & 100.0$\pm$0.0 & 100.0$\pm$0.0 & 0.0$\pm$0.0 & 0.0$\pm$0.0 & 89.3$\pm$0.6 & 80.7$\pm$1.4 & 0.0$\pm$0.0 \\

    \midrule

    \textbf{TREC} & & & & & & \\
    SNLI & 99.7$\pm$0.0 & 0.0$\pm$0.0 & 31.0$\pm$0.1 & 99.7$\pm$0.0 & 0.0$\pm$0.0 & 0.0$\pm$0.0 & 0.7$\pm$0.3 & 0.0$\pm$0.0 & 0.0$\pm$0.0 \\
    Multi30k & 100.0$\pm$0.0 & 0.0$\pm$0.0 & 14.2$\pm$0.0 & 100.0$\pm$0.0 & 0.0$\pm$0.0 & 0.0$\pm$0.0 & 0.8$\pm$0.5 & 0.0$\pm$0.0 & 0.0$\pm$0.0 \\
    WMT16 & 89.2$\pm$0.0 & 0.0$\pm$0.0 & 27.3$\pm$0.0 & 89.3$\pm$0.0 & 0.0$\pm$0.0 & 0.0$\pm$0.0 & 0.8$\pm$0.7 & 0.0$\pm$0.0 & 0.0$\pm$0.0 \\

    \bottomrule
  \end{tabular}
\end{table*}

\begin{table*}[t]
  \caption{Test accuracy (\(\uparrow\)) / ECE (\(\downarrow\)) of more sophisticated base models, averaged over five prediction runs.}
  \label{tab:acc_ece_aux}

  \centering
  \small
  \renewcommand{\tabcolsep}{10pt}

  \begin{tabular}{llll}
    \toprule
    \textbf{\footnotesize} & {\footnotesize\bf CIFAR-10} & {\footnotesize\bf CIFAR-100} \\
    \midrule

    Flipout & 91.3$\pm$0.0\,/\,10.9$\pm$0.2 & 70.4$\pm$0.1\,/\,19.8$\pm$0.2 \\
    +NC & 89.7$\pm$0.1\,/\,8.2$\pm$0.2 & 67.1$\pm$0.1\,/\,13.8$\pm$0.1 \\

    \midrule

    CSGHMC & 93.9$\pm$0.0\,/\,1.7$\pm$0.0 & 74.0$\pm$0.0\,/\,4.0$\pm$0.0 \\
    +NC & 92.2$\pm$0.0\,/\,6.2$\pm$0.0 & 71.6$\pm$0.0\,/\,2.4$\pm$0.0 \\

    \midrule

    DE & 95.7$\pm$0.0\,/\,4.5$\pm$0.0 & 80.0$\pm$0.0\,/\,1.9$\pm$0.0 \\
    +NC & 94.9$\pm$0.0\,/\,4.8$\pm$0.0 & 79.0$\pm$0.0\,/\,1.7$\pm$0.0 \\

    \bottomrule
  \end{tabular}
\end{table*}

\begin{table*}[ht]
  \caption{OOD data detection with more sophisticated base models. Values are FPR95, averaged over five prediction runs.}
  \label{tab:ood_aux_more}

  \centering

  \begin{tabular}{lrrrrrr}
    \toprule

    & \multicolumn{2}{c}{\bf Flipout} & \multicolumn{2}{c}{\bf CSGHMC} & \multicolumn{2}{c}{\bf DE} \\

    \cmidrule(r){2-3} \cmidrule(l){4-5} \cmidrule(l){6-7}

    \textbf{Datasets} & {\bf Plain} & {\bf NC} & {\bf Plain} & {\bf NC} & {\bf Plain} & {\bf NC} \\

    \midrule

    \textbf{CIFAR-10} & & & & & & \\
    SVHN & 72.1$\pm$0.3 & 39.6$\pm$0.2 & 56.8$\pm$0.0 & 16.4$\pm$0.0 & 33.9$\pm$0.0 & 8.1$\pm$0.0 \\
    LSUN-CR & 63.7$\pm$1.0 & 37.5$\pm$0.2 & 56.7$\pm$0.0 & 24.0$\pm$0.0 & 44.0$\pm$0.0 & 18.3$\pm$0.0 \\
    CIFAR-100 & 74.5$\pm$0.2 & 70.4$\pm$0.1 & 63.4$\pm$0.0 & 63.1$\pm$0.0 & 52.5$\pm$0.0 & 51.7$\pm$0.0 \\
    FMNIST-3D & 65.0$\pm$0.2 & 38.2$\pm$0.1 & 51.0$\pm$0.0 & 14.8$\pm$0.0 & 30.7$\pm$0.0 & 10.3$\pm$0.0 \\
    Uniform & 53.8$\pm$0.6 & 0.0$\pm$0.0 & 87.0$\pm$0.0 & 0.0$\pm$0.0 & 0.0$\pm$0.0 & 0.0$\pm$0.0 \\
    Smooth & 61.1$\pm$0.2 & 59.6$\pm$0.2 & 47.1$\pm$0.0 & 31.9$\pm$0.0 & 32.9$\pm$0.0 & 13.6$\pm$0.0 \\

    \midrule

    \textbf{CIFAR-100} & & & & & & \\
    LSUN-CR & 85.8$\pm$0.6 & 55.6$\pm$0.4 & 79.3$\pm$0.0 & 38.0$\pm$0.0 & 75.3$\pm$0.0 & 54.0$\pm$0.0 \\
    CIFAR-10 & 86.1$\pm$0.3 & 87.0$\pm$0.2 & 82.1$\pm$0.0 & 84.2$\pm$0.0 & 76.4$\pm$0.0 & 78.5$\pm$0.0 \\
    FMNIST-3D & 73.4$\pm$0.4 & 65.8$\pm$0.2 & 67.0$\pm$0.0 & 45.5$\pm$0.0 & 61.8$\pm$0.0 & 50.0$\pm$0.0 \\
    Uniform & 99.7$\pm$0.0 & 0.0$\pm$0.0 & 93.8$\pm$0.0 & 0.0$\pm$0.0 & 94.3$\pm$0.0 & 0.0$\pm$0.0 \\
    Smooth & 82.0$\pm$0.3 & 72.5$\pm$0.3 & 83.0$\pm$0.0 & 47.2$\pm$0.0 & 58.7$\pm$0.0 & 39.3$\pm$0.0 \\

    \bottomrule
  \end{tabular}
\end{table*}

  \end{appendices}

\end{document}